\documentclass[11pt]{article}

\usepackage[final]{acl}

\usepackage{times}
\usepackage{latexsym}

\usepackage[T1]{fontenc}

\usepackage[utf8]{inputenc}

\usepackage{microtype}

\usepackage{inconsolata}

\usepackage{graphicx}
\usepackage{makecell}
\usepackage{tabularx}
\usepackage{array}
\usepackage{makecell}
\usepackage{booktabs}
\usepackage{multirow}
\usepackage{float}
\usepackage{subfig}
\usepackage{graphicx}
\usepackage{wrapfig}
\graphicspath{{figure/}}
\usepackage[inline]{enumitem}
\usepackage{tikz}
\usepackage{xcolor}
\usepackage{tcolorbox}
\usepackage{listings}
\usepackage{pifont}
\usepackage{enumitem}
\usepackage{subcaption}
\definecolor{purple1}{HTML}{9D9CE4}
\definecolor{purple2}{HTML}{C5CCFC}
\definecolor{purple3}{HTML}{E1E4FD}
\definecolor{green1}{HTML}{00B050}
\definecolor{green2}{HTML}{8ED973}
\definecolor{green3}{HTML}{DEF4D6}
\definecolor{brown1}{HTML}{DB9679}
\definecolor{brown2}{HTML}{F2AA84}
\definecolor{brown3}{HTML}{FBE5DB}
\usetikzlibrary{matrix}
\usepackage{hyperref}
\usepackage{diagbox}
\usepackage{twemojis}
\usepackage{amssymb}  
\usepackage{tikz}
\usepackage{amssymb}
\usepackage{pifont}

\usetikzlibrary{shapes.geometric, arrows, positioning}
\usepackage{tikzscale}
\title{Business as \textit{Rule}sual: A Benchmark and Framework for Business Rule Flow Modeling with LLMs}

\author{Chen Yang\textsuperscript{1,2}, Ruping Xu\textsuperscript{1,2}, Ruizhe Li\textsuperscript{3,4}, Bin Cao\textsuperscript{1,2}\thanks{Corresponding author.}, Jing Fan\textsuperscript{1,2}\protect\footnotemark[1]\\
  Zhejiang University of Technology, China\textsuperscript{1} \\
  Zhejiang Key Laboratory of Visual Information Intelligent Processing, China \textsuperscript{2}\\
  University of Aberdeen, UK \textsuperscript{3}\\
  University of Birmingham, UK \textsuperscript{4}\\
  \texttt{\{yangchen,rupingxu,bincao,fanjing\}@zjut.edu.cn} \\
  }

\begin{document}
\maketitle
\begin{abstract}
Extracting structured procedural knowledge from unstructured business documents is a critical yet unresolved bottleneck in process automation. While prior work has focused on extracting linear action flows from instructional texts (e.g., recipes), it has insufficiently addressed the complex logical structures—such as conditional branching and parallel execution—that are pervasive in real-world regulatory and administrative documents. Furthermore, existing benchmarks are limited by simplistic schemas and shallow logical dependencies, restricting progress toward logic-aware large language models (LLMs). To bridge this ``Logic Gap'', we introduce \textbf{BREX}, a carefully curated benchmark comprising 409 real-world business documents and 2,855 expert-annotated rules. Unlike prior datasets centered on narrow service scenarios, BREX spans over 30 vertical domains, covering scientific, industrial, administrative, and financial regulations.

We further propose \textbf{ExIde}, a structure-aware reasoning framework that investigates five distinct prompting strategies, ranging from implicit semantic alignment to executable grounding via pseudo-code generation, enabling explicit modeling of rule dependencies and providing an out-of-the-box framework for different business customers without finetuning their own LLMs. We benchmark ExIde using 13 state-of-the-art LLMs. Our extensive evaluation reveals that: (1) Executable grounding serves as a superior inductive bias, significantly outperforming standard prompts in rule extraction; and (2) Reasoning-optimized models demonstrate a distinct advantage in tracing long-range dependencies and non-linear rule dependencies compared to standard instruction-tuned models. The code and dataset are available at: \url{https://github.com/oYoungCo/Business-as-Rulesual}.
\end{abstract}

\section{Introduction}

Modern enterprises rely on extensive collections of natural language regulations to govern complex operational processes, ranging from safety protocols in physics laboratories to compliance checks in  financial services \cite{kourani2024process, zhang2025unlocking}. Although these documents are written for human interpretation, their execution in real-world systems requires structured, machine-readable representations that explicitly encode logical constraints and dependencies. In practice, this translation is still performed manually by domain experts using rule engines or workflow languages (e.g., executable code or BPMN models \cite{wohed2006suitability}), making the process labor-intensive, error-prone, and difficult to maintain under frequent policy updates \cite{sivasankari2020comparative}.

We study a fundamental yet underexplored problem: \emph{how can unstructured business manuals be automatically transformed into structured rule flows that explicitly capture conditional branching, parallel constraints, and inter-rule dependencies?} We refer to this challenge as \textbf{Logic Gap}: the discrepancy between free-form natural language regulations and the executable, condition-dependent control flow required by automated systems.
\begin{figure*}[htbp]
    \centering
    \setlength{\abovecaptionskip}{0cm}
    \includegraphics[width=0.9\textwidth,
  height=0.3\textheight,
  keepaspectratio]{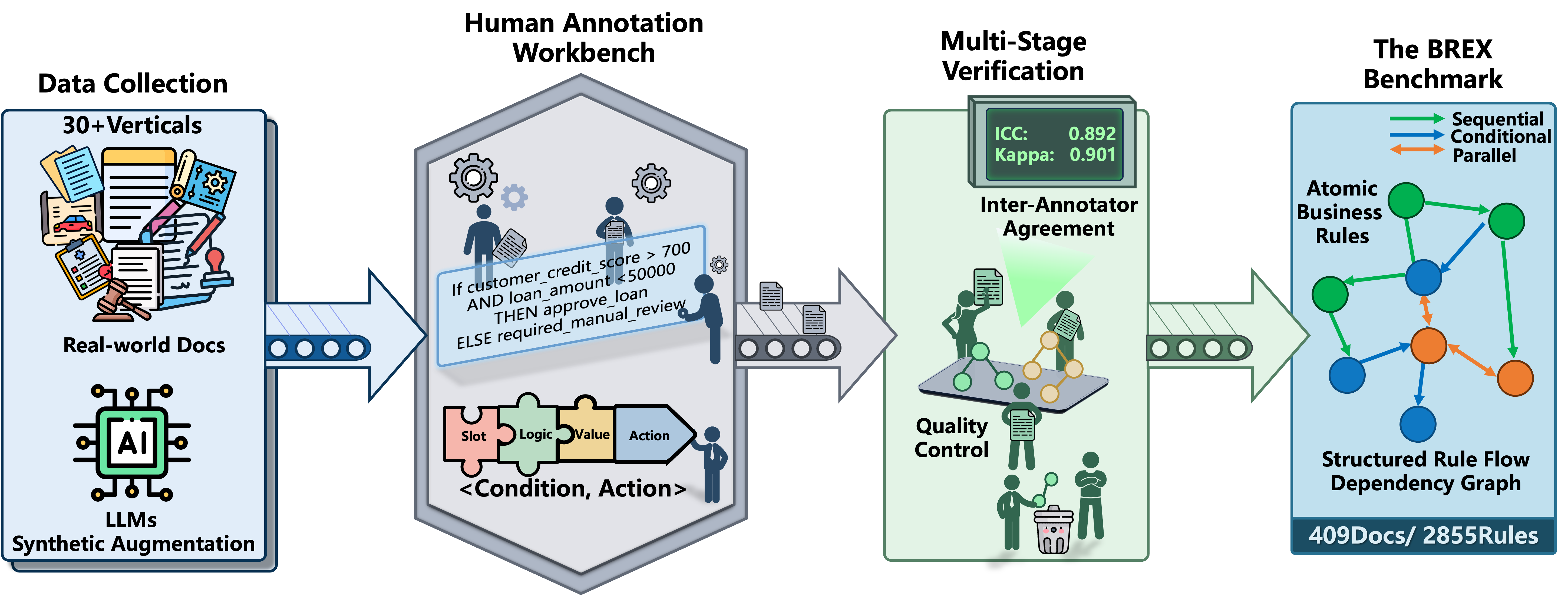}
    \caption{Construction pipeline of the BREX benchmark: 
(1) \textbf{Data collection}, combining real-world business documents with a limited amount of expert-filtered synthetic data; 
(2) \textbf{Expert annotation}, where three domain experts annotate atomic rules and their dependencies; 
and (3) \textbf{Multi-stage verification}, including quality control with another three experts and inter-annotator agreement evaluation using ICC and Kappa.}

    \label{fig:annotation}
\end{figure*}

Despite its practical importance, the Logic Gap remains poorly addressed in existing NLP benchmarks. Prior work on procedural text understanding \cite{quishpi2020extracting,DBLP:conf/acl/DuLLL24,DBLP:journals/corr/abs-2409-09191,DBLP:conf/icpm/PyrihRA25} predominantly focuses on extracting \emph{action-centric} process flows, often modeling procedures as linear or weakly structured event sequences (e.g., recipes or tutorials).  Such representations are insufficient for professional regulatory documents, where actions are governed by nested conditions, branching decisions, and parallel requirements. For example, in financial compliance, a rule such as \emph{``If the currency type is USD, select remittance type''} is activated by a condition rather than a preceding action, and subsequent checks may be triggered conditionally or executed in parallel.

This distinction is critical: \emph{action flows} describe what actions occur, whereas \emph{rule flows} specify the logical conditions under which actions are permitted or constrained. However, rule-centric modeling has been largely overlooked due to the lack of benchmarks that explicitly annotate both atomic business rules and their inter-rule dependencies at scale. As a result, existing datasets provide limited support for developing and evaluating logic-aware LLMs.

To fill this gap, we introduce \textbf{BREX} (\underline{B}usiness \underline{R}ule \underline{Ex}traction Benchmark), a dataset specifically designed for business rule flow modeling (Fig.~\ref{fig:annotation}). BREX consists of 409 real-world business documents and 2,855 expert-annotated rules spanning over 30 vertical domains, including scientific research management, industrial manufacturing, administrative approvals, and financial compliance. Each rule is formalized as a structured condition–action pair and linked via explicit dependency relations (\emph{Sequential}, \emph{Conditional}, and \emph{Parallel}), enabling systematic evaluation of both local rule extraction and global logical reasoning.

While many prior information extraction approaches rely on supervised fine-tuning, real-world deployment faces significant barriers, including strict data privacy constraints, high computational costs, and the challenge of adapting to frequent policy updates across heterogeneous formats.
As a result, there is a strong industrial demand for \textbf{out-of-the-box solutions} that can operate directly on general-purpose LLMs without task-specific retraining. Motivated by this, we propose \textbf{ExIde} designed to maximize the potential of prompt-based inference. Rather than treating rule extraction as a flat information extraction task, ExIde adopts a \emph{decompose-and-reason} strategy that introduces intermediate representations as inductive signals. We investigate five prompting strategies, ranging from implicit semantic alignment to executable grounding via pseudo-code generation, and benchmark them across 13 state-of-the-art LLMs. Our experiments reveal two key findings. First, executable grounding provides a strong inductive guidance for recovering complex business rules, consistently outperforming standard prompting strategies. Second, reasoning-optimized models exhibit a clear advantage in tracing long-range and non-linear rule dependencies, highlighting the importance of structured reasoning for logic-intensive extraction tasks.

In summary, our contributions are threefold:
\begin{itemize}[leftmargin=*, itemsep=0.1em, parsep=0pt, topsep=0.2em]
    \item We introduce BREX, a cross-domain benchmark for business rule flow modeling with expert-annotated logical conditions and dependencies across 30+ real-world domains.
    \item We propose ExIde, a structure-aware framework that leverages executable grounding to bridge natural language regulations and machine-executable rule flows, and provide out-of-the-box solutions to diverse business customers without finetuning their own LLMs.
    \item We provide a comprehensive empirical study of 13 LLMs, offering new insights into the role of executable representations and reasoning-oriented model design for logic-intensive extraction tasks.
\end{itemize}

\section{Related Work}
\label{sec:related_work}

\subsection{From Action-Centric to Rule-Centric Modeling}

Prior procedural text understanding predominantly focuses on \emph{action-centric} modeling. Early studies applied extraction techniques to instructional texts \cite{maeta2015framework,friedrich2011process}, while later work extended this to structured documents using syntactic or neural models \cite{guo2018automatic,epure2015automatic,ren2023constructing,pal2021constructing,candido2024annotated}.
However, these approaches typically conceptualize processes as linear action sequences derived from temporal order. They struggle with real-world regulations where actions are governed by complex \emph{interdependent business rules} involving conditional branching and parallel execution. This limitation is exacerbated by the lack of benchmarks explicitly annotating rule-to-rule dependencies.
A detailed comparison highlighting these differences in modeling paradigms and data sources is provided in Table \ref{tab:benchmark_comparison} in Appendix \ref{sec:benchmark_comparison}.

\subsection{LLMs for Process Understanding}

Recent work has investigated LLMs for process modeling but remains limited in logical depth. While early studies extracted linear control flows \cite{bellan2022extracting}, large-scale benchmarks like PAGED \cite{DBLP:conf/acl/DuLLL24} rely on synthetic data-to-text generation, often lacking the linguistic ambiguity and logical complexity of authentic regulatory documents. Although some approaches integrate formal constraints or instruction tuning \cite{DBLP:conf/icpm/PyrihRA25}, they typically operate under predefined schemas or focus on action-level semantics rather than atomic business rules.
In contrast, we focus on \emph{business rule flow modeling}, shifting from action sequences to jointly extracting atomic condition–action rules and reconstructing their global dependency structures. This enables a rigorous evaluation of LLMs' ability to capture the non-linear logic pervasive in real-world enterprises.

\begin{table*}[t]
    \renewcommand{\arraystretch}{1} 
    \centering
    \small 
    \scalebox{0.95}{\begin{tabular}{p{80pt} p{90pt} p{110pt} p{110pt}} \toprule[1.2pt]
    \multicolumn{4}{p{420pt}}{Our bank supports up to 39 \begin{tikzpicture}[baseline=(X.base)]
\node[fill=red!20, inner sep=2pt, rounded corners] (X) {currency types};\end{tikzpicture} of popular countries or regions around the world, \begin{tikzpicture}[baseline=(X.base)]
\node[fill=green!20, inner sep=2pt, rounded corners] (X) {including};\end{tikzpicture} \begin{tikzpicture}[baseline=(X.base)]
\node[fill=yellow!30, inner sep=2pt, rounded corners] (X) {RMB, USD, JPY, GBP, and HKD};\end{tikzpicture}. After selecting the appropriate currency, the customer needs to \begin{tikzpicture}[baseline=(X.base)]
\node[fill=blue!20, inner sep=2pt, rounded corners] (X) {choose the corresponding cash or remittance type};\end{tikzpicture} based on the type of business to be conducted thereafter. \begin{tikzpicture}[baseline=(X.base)]
\node[fill=red!20, inner sep=2pt, rounded corners] (X) {The cash or remittance type};\end{tikzpicture} \begin{tikzpicture}[baseline=(X.base)]
\node[fill=green!20, inner sep=2pt, rounded corners] (X) {includes};\end{tikzpicture} \begin{tikzpicture}[baseline=(X.base)]
\node[fill=yellow!30, inner sep=2pt, rounded corners] (X) {cash and remittance};\end{tikzpicture}. Upon completing the cash/remittance type selection, \begin{tikzpicture}[baseline=(X.base)]
\node[fill=blue!20, inner sep=2pt, rounded corners] (X) {the customer needs to provide the purchase amount};\end{tikzpicture}...} \\ \midrule
    \begin{tikzpicture}[baseline=(X.base)]
\node[fill=red!20, inner sep=2pt, rounded corners] (X) {\textbf{Slot Type}};\end{tikzpicture}  & \begin{tikzpicture}[baseline=(X.base)]
\node[fill=green!20, inner sep=2pt, rounded corners] (X) {\textbf{Logical Operator}};\end{tikzpicture}  & \begin{tikzpicture}[baseline=(X.base)]
\node[fill=yellow!30, inner sep=2pt, rounded corners] (X) {\textbf{Reference Value}};\end{tikzpicture} & \begin{tikzpicture}[baseline=(X.base)]
\node[fill=blue!20, inner sep=2pt, rounded corners] (X) {\textbf{Action}};\end{tikzpicture}  \\ \midrule
    currency types  & including  & RMB, USD, JPY, GBP, and HKD. & choose the corresponding cash or remittance type  \\ 
    The cash or remittance type  & includes  & cash and remittance  & the customer needs to provide the purchase amount  \\       \bottomrule[1.2pt]                                                       
    \end{tabular}}
    \caption{An example of business rule annotation. The model is required to extract each condition and action as a structured tuple $\langle$$\langle$Slot Type, Logical Operator, Reference Value$\rangle$, Action$\rangle$ and infer the corresponding rule dependencies. \textbf{Colors} denote corresponding components in the text.}
    \label{tab:an annotation example}
\end{table*}
\section{BREX Dataset} \label{sec:dataset}

We introduce \textbf{BREX}, a benchmark designed for \emph{business rule flow modeling}, which requires jointly extracting atomic business rules and recovering their logical dependencies from unstructured text.
Given a business document, the task is to identify a set of condition--action rules and reconstruct a dependency graph.

\subsection{Task Formalization and Annotation Schema}

Each business rule is formalized as an atomic condition--action pair.
The condition is represented as a structured triple $\langle$Slot Type, Logical Operator, Reference Value$\rangle$, while the action specifies the operation triggered when the condition is satisfied. Formal definitions are provided in Appendix~\ref{sec:definition}, and an annotated example is illustrated in Table~\ref{tab:an annotation example}. Complex rules involving multiple constraints are normalized into atomic units and linked through explicit dependency relations, enabling compositional modeling of non-linear logic.

We consider three types of dependency relationships:
\emph{Sequential}, where one rule must be executed before another;
\emph{Conditional}, where different outcomes of a rule trigger different subsequent rules;
and \emph{Parallel}, where multiple rules must be executed concurrently. Illustrative examples of these dependencies are provided in Appendix~\ref{sec:dependency relationship}.

\subsection{Dataset Construction}
To ensure both coverage and realism, we combine publicly available real-world documents with a limited amount of synthetic data generated using Gemini 2.5 Pro \cite{comanici2025gemini}. Importantly, synthetic texts are used only to augment underrepresented logical structures (e.g., nested conditions and parallel constraints) and are strictly filtered and revised by domain experts to ensure consistency with real-world regulatory language. The construction process (Fig~\ref{fig:annotation}) involves:

\begin{enumerate}[leftmargin=*, itemsep=0.1em, parsep=0pt, topsep=0.2em]
    \item \textbf{Data Collection}: We gathered business texts from over \textbf{30 distinct vertical domains}. Crucially, unlike prior datasets limited to simple service interactions, BREX covers: Scientific \& Industrial Logic, Administrative \& Legal Logic, Service \& Transactional Logic, etc.
    Synthetic texts were generated to augment these specific domains. The generation prompt is shown in Fig \ref{fig:data generation prompt} in Appendix.
   
    \item \textbf{Expert Annotation}: Three domain experts annotated atomic rules and their dependency relations following a unified schema.
    
    \item \textbf{Multi-Stage Verification}: To ensure the quality and consistency of the annotations, a separate team of three experts reviewed all annotations. Any ambiguous rules were discarded or re-annotated to ensure the dataset serves as a gold-standard benchmark.
\end{enumerate}

The BREX dataset contains \textbf{409 documents} and \textbf{2,855 business rules}, with an average of \textbf{7 rules per document}. See Table~\ref{tab:statistical information} in Appendix for more statistics. Notably, while sequential dependencies are common, over 30\% of rules participate in conditional or parallel relations, highlighting the prevalence of non-linear logic in real-world regulations.

\subsection{Dataset Analysis}
We assessed the quality of BREX from two perspectives: the linguistic quality of the texts and the reliability of the logical annotations.


\textbf{Text Quality Assessment}: Following prior studies \cite{miller1979humanistic, DBLP:conf/acl/DuLLL24}, three annotators rated the business texts on Readability, Accuracy, Clarity, Simplicity, and Usability (Scale 1-5) (See Appendix~\ref{sec:text_quality_assessment_criteria}). As shown in Fig~\ref{ICC} in Appendix, the texts received high ratings across all dimensions. We computed the Intraclass Correlation Coefficient (ICC) \cite{shrout1979intraclass} to measure agreement. The average ICC was \textbf{0.892}, indicating excellent consistency and high-quality textual data.

\textbf{Inter-Annotator Agreement (IAA)}: Evaluating IAA for structured rule extraction is challenging as it involves both span identification and relation classification. To our knowledge, there is no established metric for directly measuring inter-annotator agreement on structured rule dependency graphs. To address this, we adopted a proxy evaluation method by projecting the rule annotations into a Named Entity Recognition (NER) format (as shown in Table~\ref{tab:ner_annotation_example} in \ref{sec:iaa_details}). Three NLP experts annotated a subset of data using BIO tagging for Slot Types, Reference Values and Action. We then computed Fleiss' Kappa \cite{artstein2017inter}. The resulting score of \textbf{0.901} indicates \textit{almost perfect agreement}, confirming that our definitions of business rules are unambiguous and the annotations are highly reliable.

\section{Methodology} \label{sec:methodology}

\begin{figure*}[htbp]
    \centering
    \includegraphics[width=0.9\textwidth,
  height=0.3\textheight,
  keepaspectratio]{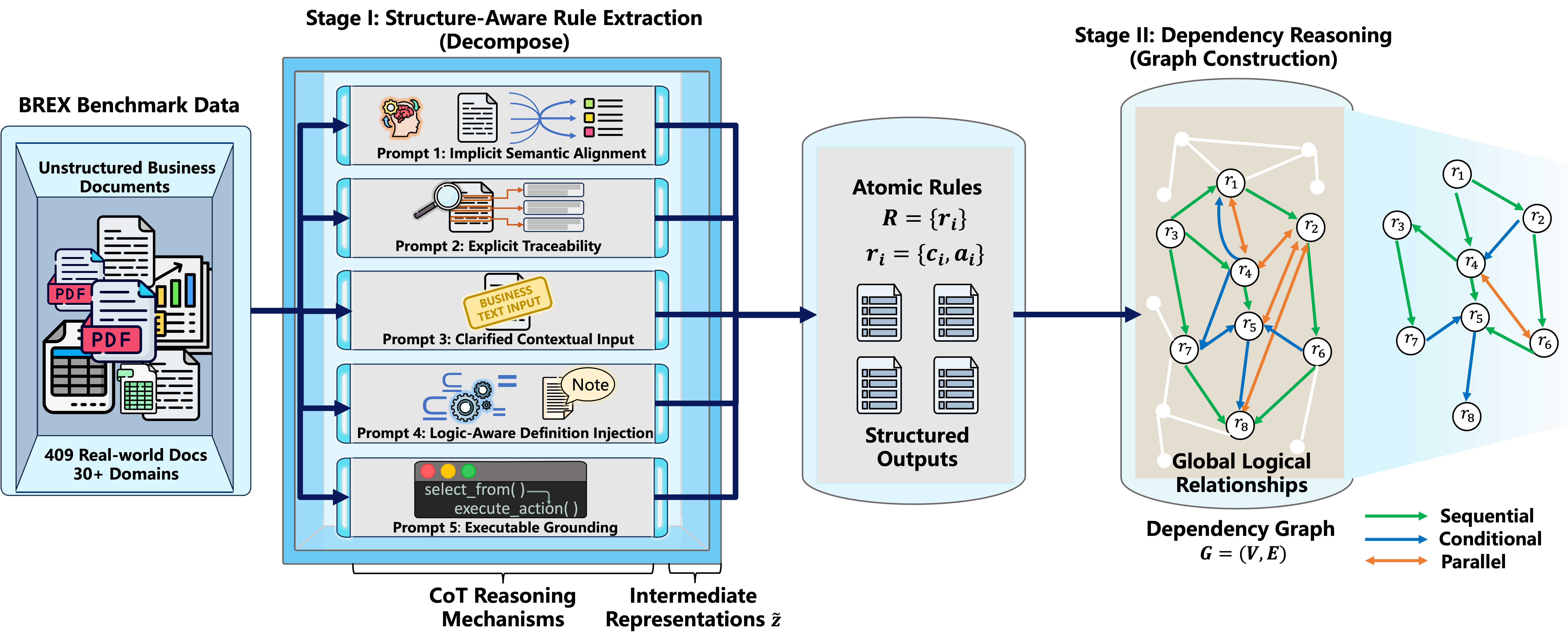}
    \caption{Overview of the \textbf{ExIde} framework. It adopts a \textit{decompose-and-reason} strategy: (1) \textbf{Structure-Aware Extraction}, which employs five distinct reasoning mechanisms (e.g., executable grounding) to extract atomic rules; and (2) \textbf{Dependency Reasoning}, which infers global logical relationships (Sequential, Conditional, Parallel) among the extracted rules.}
    \label{fig:framework}
\end{figure*}

We propose \textbf{ExIde}, a structure-aware framework for \emph{business rule flow modeling}. Given a business document $x$, ExIde aims to recover (i) a set of atomic rules $R=\{r_i\}$ and (ii) a dependency graph $G$ that encodes \textit{Sequential}, \textit{Conditional}, and \textit{Parallel} relations among rules. As illustrated in Fig.~\ref{fig:framework}, ExIde adopts a \emph{decompose-and-reason} strategy that separates rule extraction from global dependency reasoning.

\subsection{Stage I: Structure-Aware Rule Extraction}
In the first stage, ExIde extracts atomic rules of the form $r_i=(c_i,a_i)$, where each condition
$c_i=\langle s_i, j_i, v_i\rangle$ follows the schema defined in Section~\ref{sec:dataset}, and $a_i$ denotes the corresponding action. Rather than treating rule extraction as a flat information extraction task, ExIde introduces \emph{intermediate reasoning structures} to guide LLMs toward logic-consistent outputs.

To study the effect of different inductive biases, we design five prompting strategies (P1--P5) that share the same output schema but differ in how intermediate reasoning is encouraged.
These strategies range from implicit semantic alignment to explicit executable grounding.
Specifically, Prompt~5 introduces an intermediate pseudo-code representation, in which the business logic is first expressed using simple procedural primitives (e.g., conditional branches) before structured rules are extracted. This design encourages early resolution of nested conditions and control flow, providing a structural scaffold for logic-intensive extraction.

All prompting strategies adopt a chain-of-thought style reasoning format \cite{wei2022chain} (see Fig. \ref{fig:5 prompt template}), but only the final structured outputs are used for evaluation. The specific reasoning mechanisms are as follows:

\begin{itemize}[leftmargin=*, itemsep=0.1em, parsep=0pt, topsep=0.2em]
    \item \textbf{Prompt 1: Implicit Semantic Alignment.} 
    This prompt (Fig~\ref{fig:prompt1}) establishes a baseline reasoning path. It requires the LLMs to first generate a natural language explanation (\textbf{Explain} section) that summarizes the business logic before populating the structured \textbf{Output}. The mapping between the source text and the extracted rules is implicit, relying on the model's internal attention mechanism to align semantics with the schema. 

    \item \textbf{Prompt 2: Explicit Traceability (Alignment).} 
    To mitigate hallucinations and enhance interpretability, this strategy (Fig~\ref{fig:prompt2}) enforces a strict one-to-one mapping. The model must explicitly link each explanatory sentence to its corresponding business rule using declarative phrasing (e.g., ``\textit{This sentence corresponds to the business rule: ...}''). By forcing explicit alignment, we aim to reduce the generation of unsupported rules.

    \item \textbf{Prompt 3: Clarified Contextual Input.} 
    Building on Prompt 1, this variant refines the input by explicitly labeling the ``\textbf{Input}'' field as ``\textbf{Business Text Input}''. While a subtle modification, this aims to prime the model's domain awareness, ensuring the input is processed strictly as regulatory text rather than general prose.

    \item \textbf{Prompt 4: Logic-Aware Definition Injection.} 
    Complex logical operators (e.g., distinguishing \textit{contains} vs. \textit{equal to}) are frequent sources of error. This prompt (Fig~\ref{fig:prompt3}) injects detailed constraints into the context, specifically guiding the handling of multi-value slot types. It introduces a dedicated ``Note'' section to disambiguate set inclusion from equality, enforcing rigorous adherence to the logical definitions in Section~\ref{sec:definition}. 

    \item \textbf{Prompt 5: Executable Grounding (Pseudo-Code).} 
    Prompt~5 (Fig~\ref{fig:prompt5}) introduces an intermediate pseudo-code representation $\tilde{z}$ before producing the final rules. The model first translates $x$ into pseudo-code using a small set of primitives (e.g., \textit{select\_from()}, \textit{execute\_action()}), and then extracts $R$ from $\tilde{z}$. This design encourages early resolution of nested conditions and control-flow, leveraging code-like structure as an inductive bias for logic-intensive extraction.
\end{itemize}

Prompts 3, 4, and 5 represent targeted modifications to the baseline (Prompt 1), allowing us to systematically evaluate the impact of context clarification, logic definition, and code grounding on extraction performance.


\subsection{Stage II: Dependency Graph Reconstruction}
In the second stage, ExIde reconstructs a global dependency graph $G=(V,E)$ over the extracted rules. 
Each node $v_i \in V$ corresponds to a rule $r_i$, and each directed edge $e_{ij} \in E$ is labeled as \emph{Sequential}, \emph{Conditional}, \emph{Parallel}, or \emph{None}. Instead of performing independent pairwise classification, which is prone to generating contradictory edges, we propose a \textbf{Listwise Contextual Reasoning} approach. We employ a dedicated dependency reasoning prompt (Fig. \ref{fig:prompt dependency relationship}) that feeds the full business text alongside all candidate rule pairs into the model within a single prompt. This allows the model to globally resolve dependencies and maintain structural coherence when assembling the typed adjacency matrix. Furthermore, while the number of candidate pairs technically scales quadratically $\mathcal{O}(N^2)$, real-world business documents typically contain a manageable number of rules (e.g., an average of 7 rules per document in our dataset). Consequently, processing all pairs in a single listwise prompt is highly computationally feasible and significantly reduces the inference overhead compared to making separate calls. This global reasoning enables ExIde to successfully recover complex, non-linear rule flow structures.

\section{Experiments}

In this section, we present a comprehensive evaluation of the \textbf{ExIde} framework. Our evaluation focuses on three questions:
(1) Does executable grounding provide a stronger inductive signal for business rule extraction?
(2) Are reasoning-oriented models necessary for recovering global rule dependencies?
(3) How robust is ExIde under increasing logical complexity?

\begin{table*}[t]
    \centering
    \setlength{\tabcolsep}{23pt}
    \renewcommand{\arraystretch}{0.9}
    \scalebox{0.7}{
\begin{tabular}{c|ccccc|c} \toprule[1.5pt]
        \textbf{Model} & \textbf{\makecell{P1}}        & \textbf{\makecell{P2}} & \textbf{\makecell{P3}} & \textbf{\makecell{P4}} & \textbf{\makecell{P5}} & \textbf{Avg}   \\ \hline
        \multicolumn{7}{c}{\textbf{Closed source}}                                                                \\ \hline
Gemini-2.5-flash    & 0.888                                                                      & 0.845                         & 0.873                                                                     & \begin{tikzpicture}[baseline=(X.base)]
\node[fill=purple3, inner sep=5pt, rounded corners] (X) {0.891};\end{tikzpicture}                                               & 0.875                         & 0.874                         \\
Gemini-2.5-pro      & \begin{tikzpicture}[baseline=(X.base)]
\node[fill=purple1, inner sep=5pt, rounded corners] (X) {0.916};\end{tikzpicture}                                              & \begin{tikzpicture}[baseline=(X.base)]
\node[fill=purple1, inner sep=5pt, rounded corners] (X) {0.913};\end{tikzpicture} & \begin{tikzpicture}[baseline=(X.base)]
\node[fill=purple1, inner sep=5pt, rounded corners] (X) {0.907};\end{tikzpicture}                                             & \begin{tikzpicture}[baseline=(X.base)]
\node[fill=purple2, inner sep=5pt, rounded corners] (X) {0.892};\end{tikzpicture}                                                & 0.899                         & \begin{tikzpicture}[baseline=(X.base)]
\node[fill=green1, font=\bfseries, inner sep=5pt, rounded corners] (X) {0.905};\end{tikzpicture} \\
GPT-5-mini          & 0.794                                                                      & 0.811                         & 0.789                                                                     & 0.824                                                                        & 0.806                         & 0.809                         \\
GPT-5               & 0.876                                                                      & 0.88                          & 0.877                                                                     & 0.802                                                                        & 0.873                         & 0.862                         \\
\hline
\multicolumn{7}{c}{\textbf{Open source}}                                                                                                                                                                                                                                                                                                                             \\ \hline
Kimi-k2-Instruct    & \begin{tikzpicture}[baseline=(X.base)]
\node[fill=purple2, inner sep=5pt, rounded corners] (X) {0.908};\end{tikzpicture}                                             & \begin{tikzpicture}[baseline=(X.base)]
\node[fill=purple3, inner sep=5pt, rounded corners] (X) {0.905};\end{tikzpicture} & \begin{tikzpicture}[baseline=(X.base)]
\node[fill=purple2, inner sep=5pt, rounded corners] (X) {0.899};\end{tikzpicture}                                             & 0.883                                                                        & \begin{tikzpicture}[baseline=(X.base)]
\node[fill=purple1, inner sep=5pt, rounded corners] (X) {0.915};\end{tikzpicture} & \begin{tikzpicture}[baseline=(X.base)]
\node[fill=green2, inner sep=5pt, rounded corners] (X) {0.902};\end{tikzpicture} \\
Kimi-k2-Thinking    & 0.892                                                                      & \begin{tikzpicture}[baseline=(X.base)]
\node[fill=purple3, inner sep=5pt, rounded corners] (X) {0.905};\end{tikzpicture} & 0.889                                                                     & 0.884                                                                        & \begin{tikzpicture}[baseline=(X.base)]
\node[fill=purple3, inner sep=5pt, rounded corners] (X) {0.902};\end{tikzpicture} & \begin{tikzpicture}[baseline=(X.base)]
\node[fill=green3, inner sep=5pt, rounded corners] (X) {0.894};\end{tikzpicture} \\
DeepSeek-3.1        & 0.883                                                                      & 0.896                         & 0.879                                                                     & 0.884                                                                        & 0.896                         & 0.888                         \\
DeepSeek-3.2-exp    & \begin{tikzpicture}[baseline=(X.base)]
\node[fill=purple3, inner sep=5pt, rounded corners] (X) {0.902};\end{tikzpicture}                                              & \begin{tikzpicture}[baseline=(X.base)]
\node[fill=purple2, inner sep=5pt, rounded corners] (X) {0.909};\end{tikzpicture} & \begin{tikzpicture}[baseline=(X.base)]
\node[fill=purple3, inner sep=5pt, rounded corners] (X) {0.898};\end{tikzpicture}                                             & \begin{tikzpicture}[baseline=(X.base)]
\node[fill=purple1, inner sep=5pt, rounded corners] (X) {0.895};\end{tikzpicture}                                                & \begin{tikzpicture}[baseline=(X.base)]
\node[fill=purple2, inner sep=5pt, rounded corners] (X) {0.905};\end{tikzpicture} & \begin{tikzpicture}[baseline=(X.base)]
\node[fill=green2, inner sep=5pt, rounded corners] (X) {0.902};\end{tikzpicture} \\
DeepSeek-R1         & 0.885                                                                      & 0.867                         & 0.876                                                                     & 0.875                                                                        & 0.875                         & 0.876                         \\
Qwen3-30B-Instruct  & 0.885                                                                      & 0.886                         & 0.879                                                                     & 0.870                                                                         & 0.880                          & 0.880                          \\
Qwen3-30B-Thinking  & 0.810                                                                       & 0.839                         & 0.805                                                                     & 0.859                                                                        & 0.873                         & 0.837                         \\
Qwen3-235B-Instruct & 0.898                                                                      & 0.888                         & 0.895                                                                     & 0.888                                                                        & 0.892                         & 0.892                         \\
Qwen3-235B-Thinking & 0.885                                                                      & 0.877                         & 0.878                                                                     & 0.880                                                                         & 0.881                         & 0.880                         \\ \hline
\textbf{Avg}                 & \begin{tikzpicture}[baseline=(X.base)]
\node[fill=brown2, inner sep=5pt, rounded corners] (X) {0.879};\end{tikzpicture}                                              & \begin{tikzpicture}[baseline=(X.base)]
\node[fill=brown2, inner sep=5pt, rounded corners] (X) {0.879};\end{tikzpicture} & \begin{tikzpicture}[baseline=(X.base)]
\node[fill=brown3, inner sep=5pt, rounded corners] (X) {0.873};\end{tikzpicture}                                            & 0.871                                                                        & \begin{tikzpicture}[baseline=(X.base)]
\node[fill=brown1, font=\bfseries,inner sep=5pt, rounded corners] (X) {0.882};\end{tikzpicture} &  \\ \bottomrule[1.5pt]                            
\end{tabular}}
\caption{\label{tab:ner_results} Rule extraction performance (NER F1) across five prompting strategies. 
\textbf{P1--P5} correspond to five prompting strategies, respectively.
\textbf{Avg} denotes the average F1 score over all models or over all promptings. \textbf{Darker} shades indicate higher performance.}
\end{table*}

\subsection{Experimental Setup}

\paragraph{Models.}
We conduct a comprehensive evaluation across 13 LLMs, spanning both closed-source and open-source architectures, across 5 model families: GPT, Gemini, Kimi, DeepSeek, and Qwen, which cover various model sizes. 
For Stage I, we assess structured rule extraction using NER F1 for \emph{Slot Type}, \emph{Reference Value}, and \emph{Action}, and F1 for \emph{Logical Operator} classification.
For Stage II, dependency identification is formulated as a multi-class classification problem over \emph{Sequential}, \emph{Conditional}, \emph{Parallel}, and \emph{None} relations, evaluated using F1 score.



\subsection{Stage I Results: Executable Grounding as a Structural Scaffold}
Table~\ref{tab:ner_results} and Table~\ref{tab:main_results1} report the performance of five prompting strategies across all models.
A consistent trend emerges: \textbf{Executable Grounding (P5)} achieves the strongest performance, with the highest average F1 scores for both NER and logical operator classification.

This result supports our hypothesis that introducing an intermediate executable representation provides an effective inductive signal for logic-intensive extraction.
By forcing early resolution of conditional branches and control flow, pseudo-code grounding helps models produce more precise and structurally consistent outputs.

In contrast, \textbf{Logic-Aware Definition Injection (P4)} performs worst on average.
Despite explicitly providing logical definitions, P4 degrades performance under dense inputs, suggesting that verbose declarative constraints increase prompt-level reasoning overhead and interfere with the model's ability to maintain consistent rule instantiation under dense inputs. These results indicate that procedural representations are more effective than textual definitions for guiding logical reasoning.


\begin{table*}[t]
    \centering
    \setlength{\tabcolsep}{23pt}
    \renewcommand{\arraystretch}{0.9}
    \scalebox{0.7}{
\begin{tabular}{c|ccccc|c} \toprule[1.5pt]
        \textbf{Model} & \textbf{\makecell{P1}}        & \textbf{\makecell{P2}} & \textbf{\makecell{P3}} & \textbf{\makecell{P4}} & \textbf{\makecell{P5}} & \textbf{Avg}   \\ \hline
        \multicolumn{7}{c}{\textbf{Closed source}}                                                                \\ \hline
Gemini-2.5-flash    & \begin{tikzpicture}[baseline=(X.base)]
\node[fill=purple2, inner sep=5pt, rounded corners] (X) {0.892};\end{tikzpicture}                                              & 0.818                         & 0.864                                                                     & \begin{tikzpicture}[baseline=(X.base)]
\node[fill=purple2, inner sep=5pt, rounded corners] (X) {0.803};\end{tikzpicture}                                                & \begin{tikzpicture}[baseline=(X.base)]
\node[fill=purple2, inner sep=5pt, rounded corners] (X) {0.896};\end{tikzpicture} & \begin{tikzpicture}[baseline=(X.base)]
\node[fill=green3, inner sep=5pt, rounded corners] (X) {0.855};\end{tikzpicture} \\
Gemini-2.5-pro      & \begin{tikzpicture}[baseline=(X.base)]
\node[fill=purple1, inner sep=5pt, rounded corners] (X) {0.914};\end{tikzpicture}                                              & 0.876                         & 0.838                                                                     & \begin{tikzpicture}[baseline=(X.base)]
\node[fill=purple1, inner sep=5pt, rounded corners] (X) {0.811};\end{tikzpicture}                                                & \begin{tikzpicture}[baseline=(X.base)]
\node[fill=purple1, inner sep=5pt, rounded corners] (X) {0.931};\end{tikzpicture}  & \begin{tikzpicture}[baseline=(X.base)]
\node[fill=green1, font=\bfseries,inner sep=5pt, rounded corners] (X) {0.874};\end{tikzpicture} \\
GPT-5-mini          & 0.793                                                                      & 0.824                         & 0.831                                                                     & 0.796                                                                        & 0.818                         & 0.812                         \\
GPT-5               & 0.838                                                                      & \begin{tikzpicture}[baseline=(X.base)]
\node[fill=purple3, inner sep=5pt, rounded corners] (X) {0.879};\end{tikzpicture} & \begin{tikzpicture}[baseline=(X.base)]
\node[fill=purple2, inner sep=5pt, rounded corners] (X) {0.880};\end{tikzpicture}                                              & 0.755                                                                        & 0.843                         & 0.839                         \\ \hline
\multicolumn{7}{c}{\textbf{Open source}}                                                                                                                                                                                                                                                                                                                            \\ \hline 
Kimi-k2-Instruct    & 0.843                                                                      & 0.876                         & 0.851                                                                     & 0.767                                                                        & 0.838                         & 0.835                         \\
Kimi-k2-Thinking    & 0.824                                                                      & \begin{tikzpicture}[baseline=(X.base)]
\node[fill=purple2, inner sep=5pt, rounded corners] (X) {0.882};\end{tikzpicture} & \begin{tikzpicture}[baseline=(X.base)]
\node[fill=purple3, inner sep=5pt, rounded corners] (X) {0.874};\end{tikzpicture}                                             & 0.777                                                                        & 0.837                         & 0.839                         \\
DeepSeek-3.1        & 0.822                                                                      & \begin{tikzpicture}[baseline=(X.base)]
\node[fill=purple2, inner sep=5pt, rounded corners] (X) {0.882};\end{tikzpicture} & 0.871                                                                     & 0.776                                                                        & 0.848                         & 0.840                          \\
DeepSeek-3.2-exp    & 0.865                                                                      & \begin{tikzpicture}[baseline=(X.base)]
\node[fill=purple1, inner sep=5pt, rounded corners] (X) {0.896};\end{tikzpicture} & \begin{tikzpicture}[baseline=(X.base)]
\node[fill=purple1, inner sep=5pt, rounded corners] (X) {0.895};\end{tikzpicture}                                             & \begin{tikzpicture}[baseline=(X.base)]
\node[fill=purple3, inner sep=5pt, rounded corners] (X) {0.800};\end{tikzpicture}                                                 & 0.848                         & \begin{tikzpicture}[baseline=(X.base)]
\node[fill=green2, inner sep=5pt, rounded corners] (X) {0.861};\end{tikzpicture} \\
DeepSeek-R1         & 0.855                                                                      & 0.865                         & 0.816                                                                     & 0.778                                                                        & 0.845                         & 0.832                         \\
Qwen3-30B-Instruct  & 0.820                                                                       & 0.834                         & 0.813                                                                     & 0.735                                                                        & 0.811                         & 0.803                         \\
Qwen3-30B-Thinking  & 0.740                                                                       & 0.720                          & 0.644                                                                     & 0.724                                                                        & 0.826                         & 0.731                         \\
Qwen3-235B-Instruct & 0.837                                                                      & 0.818                         & 0.838                                                                     & 0.771                                                                        & 0.838                         & 0.820                          \\
Qwen3-235B-Thinking & \begin{tikzpicture}[baseline=(X.base)]
\node[fill=purple3, inner sep=5pt, rounded corners] (X) {0.882};\end{tikzpicture}                                              & 0.834                         & 0.841                                                                     & 0.755                                                                        & \begin{tikzpicture}[baseline=(X.base)]
\node[fill=purple3, inner sep=5pt, rounded corners] (X) {0.867};\end{tikzpicture} & 0.836                         \\ \hline
\textbf{Avg}                 & \begin{tikzpicture}[baseline=(X.base)]
\node[fill=brown3, inner sep=5pt, rounded corners] (X) {0.840};\end{tikzpicture}                                              & \begin{tikzpicture}[baseline=(X.base)]
\node[fill=brown2, inner sep=5pt, rounded corners] (X) {0.846};\end{tikzpicture} & 0.835                                                                     & 0.773                                                                        & \begin{tikzpicture}[baseline=(X.base)]
\node[fill=brown1, font=\bfseries, inner sep=5pt, rounded corners] (X) {0.850};\end{tikzpicture}   \\ \bottomrule[1.5pt]                              
\end{tabular}}
\caption{\label{tab:main_results1}Logical operator classification performance (F1) under different prompting strategies. 
\textbf{P1--P5} follow the same prompting design as in Table~\ref{tab:ner_results}. 
\textbf{Avg} denotes the average F1 score over all models or over all promptings. }
\end{table*}
\begin{figure}[t]
    \centering
    \includegraphics[width=0.5\textwidth,
  height=0.2\textheight,
  keepaspectratio]{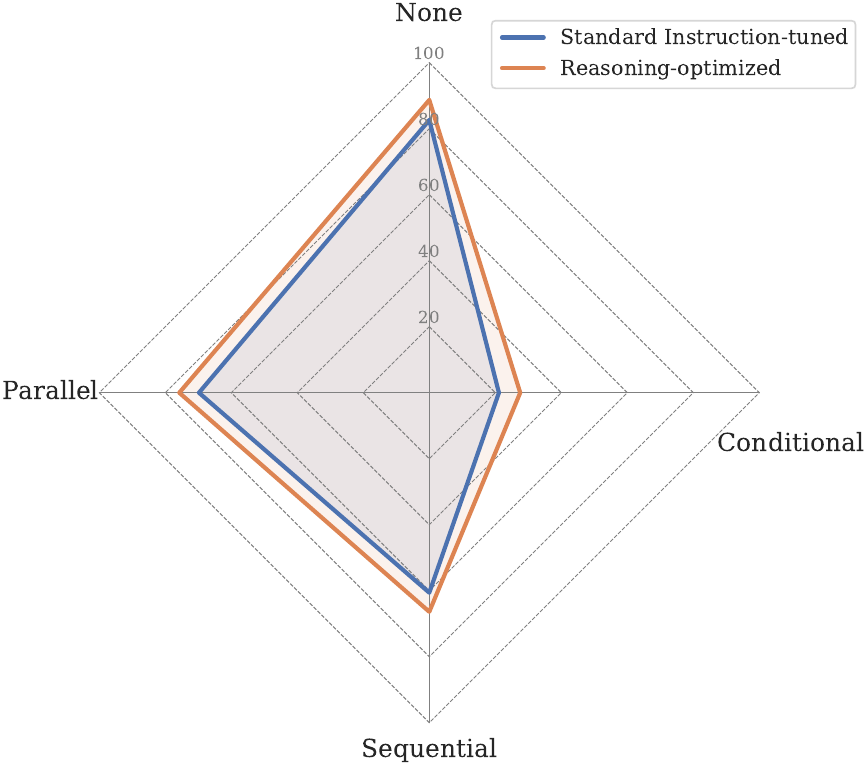}
    \caption{Performance comparison across dependency types using a radar plot.
Reasoning-optimized models consistently outperform instruction-tuned baselines.}
    \label{fig:radar dependency relationship}
\end{figure}


\subsection{Stage II Results: Reasoning Models Bridge the Logic Gap}


Table~\ref{tab:accuracy results of dependency relationships} summarizes dependency identification performance. Overall, models equipped with explicit reasoning mechanisms consistently outperform standard instruction-tuned counterparts, highlighting a clear Reasoning Gap. Dependency identification requires tracing causal relations across non-adjacent rules. While instruction-tuned models often rely on shallow semantic similarity, reasoning-oriented models better simulate execution paths, enabling accurate recovery of long-range and non-linear dependencies. This trend is further illustrated in Figure~\ref{fig:radar dependency relationship}, which provides a dependency-type–wise comparison. The performance gains of reasoning models are most pronounced on \textbf{non-linear dependencies}, particularly \textit{Conditional} and \textit{Parallel} relations, whereas improvements on linear \textit{Sequential} relations are comparatively modest. This pattern suggests that explicit reasoning mechanisms are especially effective in handling branching and concurrent logic structures, rather than purely linear execution flows. This advantage is most pronounced in smaller models: \textbf{Qwen3-30B-Thinking} (0.485) achieves a substantial \textbf{+6.0\% improvement} over its Instruct counterpart (0.425), suggesting that reasoning capabilities can effectively compensate for model size constraints.


\begin{figure*}[t]
    \centering
    \begin{minipage}[b]{0.63\linewidth}
        \centering
        \includegraphics[width=\linewidth]{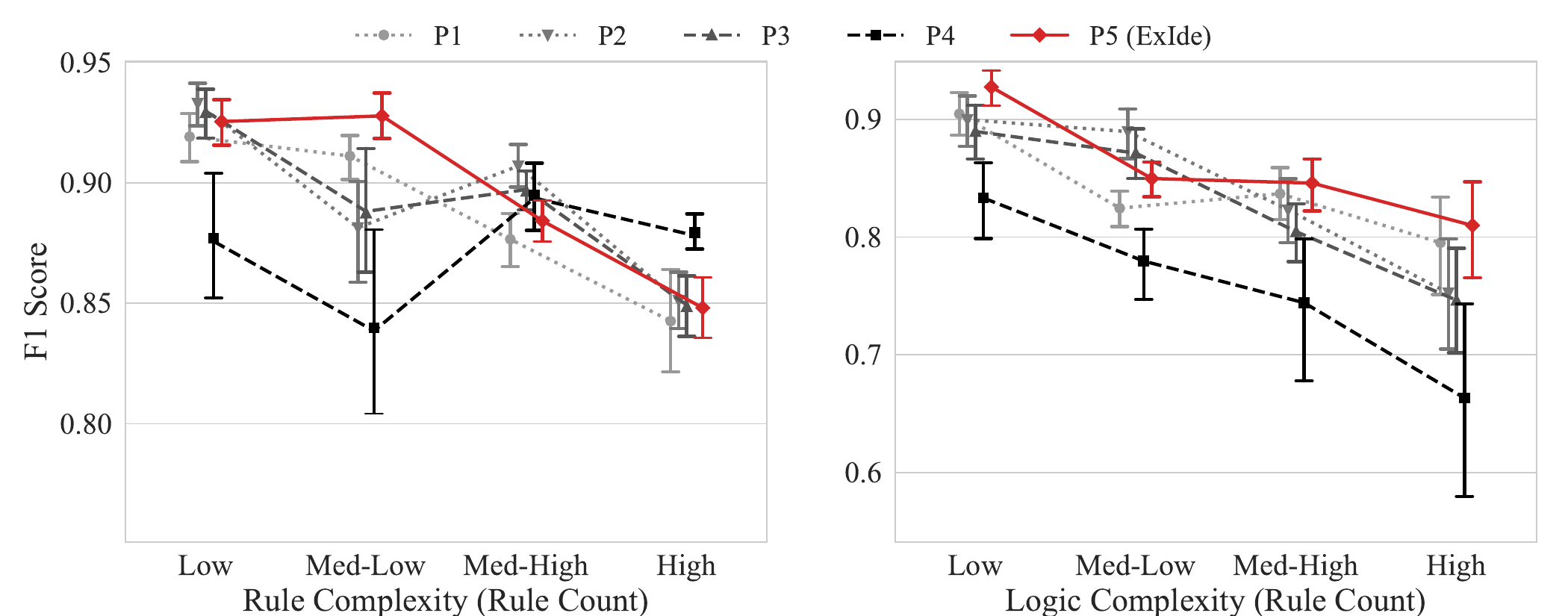}
    \end{minipage}
    \hfill 
    \begin{minipage}[b]{0.35\linewidth}
        \centering
        \includegraphics[width=\linewidth]{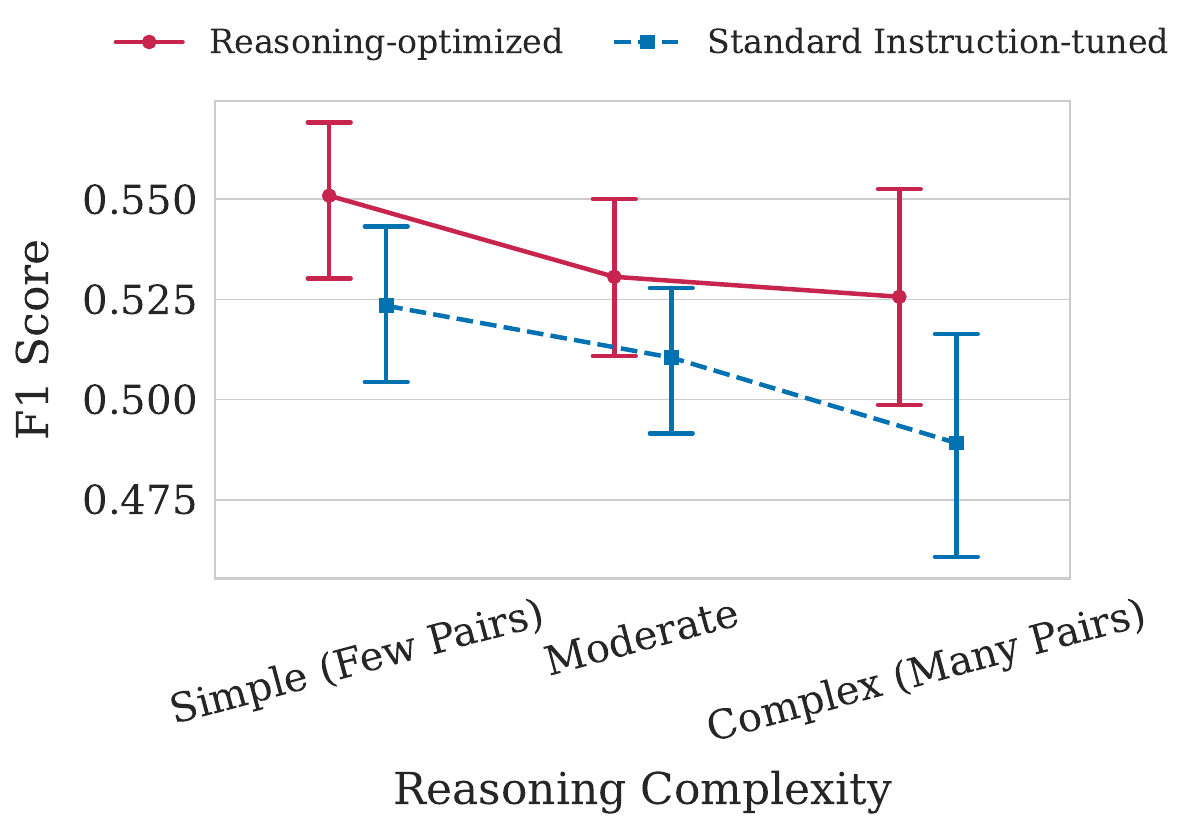}
    \end{minipage}
    
    \caption{Robustness and Complexity Analysis. The \textit{left} figures show the performance under varying rule and logic complexities, while the \textit{right} figure illustrates the impact of reasoning complexity.}
    \label{fig:robustness_complexity}
\end{figure*}

\subsection{Analysis on Complexity and Robustness}

To assess the scalability of ExIde, we analyze performance under increasing complexity dimensions: (1)~\textbf{Rule Density} (number of rules per document) for business rule extraction, and (2)~\textbf{Reasoning Complexity} (number of rule pairs) for dependency relationship identification.

We group test documents by the number of rules they contain, a proxy for information density, and analyze the performance trends of different prompting strategies in the \textit{left} panel of Figure~\ref{fig:robustness_complexity}. As rule density increases, all methods exhibit performance degradation; however, \textbf{P5} shows a relatively flat performance curve, maintaining stable performance even in high-density documents containing more than 10 rules. This robustness indicates that executable grounding serves as an effective structural anchor, mitigating the ``lost-in-the-middle'' effect \cite{he2024never} in long regulatory texts. In contrast, \textbf{P4} exhibits a heavy-tailed variance distribution in the \textit{Med-Low} bucket. Case analysis (Appendix~\ref{sec:case_study}) reveals that when business texts enumerate multiple options leading to semantically similar or identical downstream actions, models prompted by P4 frequently over-merge distinct equality conditions into a single \textit{equal} rule. This behavior violates intended atomic semantics of equality and collapses multiple rule instantiations into an abstract representation. Consequently, some models preserve fine-grained rule coverage and achieve near-perfect scores, while others suffer sharp recall degradation, resulting in observed heavy-tailed variance under string-based evaluation metrics.

Although \textbf{P5} achieves the best overall accuracy, we observe a bucket-specific reversal for a subset of models in high-density documents. This phenomenon arises from two competing effects (see
Appendix~\ref{sec:case_study}). On the one hand, the pseudo-code scaffold in P5 introduces additional prompt overhead and encourages the reification of intermediate meta-level variables, which can generate spurious rule tuples and reduce precision in densely coupled workflows. On the other hand, P4's explicit
decomposition of \emph{contain} versus \emph{equal} relations can be advantageous in texts dominated by large enumerations or threshold-based rules, as it encourages models to remain grounded in explicit surface entities rather than synthesizing abstract state variables. However, this advantage comes at the cost of increased variance. By introducing an additional decomposition stage, P4 becomes highly sensitive to action-scope interpretation and surface-form drift in action phrases, leading to inconsistent behavior across models and instances. Taken together, these
effects explain both the higher variance of P4 and its occasional performance advantage in the High-density bucket.

\begin{table}[tb]
    \centering
    \renewcommand{\arraystretch}{1}
\small
    \setlength{\tabcolsep}{20pt}
\begin{tabular}{cc} \toprule[1.5pt]
\multicolumn{1}{c}{\textbf{Model}} & \multicolumn{1}{c}{\textbf{F1}} \\ \hline
\multicolumn{2}{c}{\textbf{Closed source}}                        \\ \hline
Gemini-2.5-Flash          & \begin{tikzpicture}[baseline=(X.base)]
\node[fill=purple2, inner sep=5pt, rounded corners] (X) {0.629};\end{tikzpicture}                        \\
Gemini-2.5-Pro            & \begin{tikzpicture}[baseline=(X.base)]
\node[fill=purple1, font=\bfseries,inner sep=5pt, rounded corners] (X) {0.764};\end{tikzpicture}                        \\
GPT-5-mini                & 0.550                        \\
GPT-5                     & \begin{tikzpicture}[baseline=(X.base)]
\node[fill=purple3, inner sep=5pt, rounded corners] (X) {0.589};\end{tikzpicture}    \\ \hline
\multicolumn{2}{c}{\textbf{Open source}}                          \\ \hline
Kimi-k2-Instruct          & 0.482                        \\
Kimi-k2-Thinking          & 0.500                        \\
DeepSeek-3.1              & 0.575                        \\
DeepSeek-3.2-exp          & \begin{tikzpicture}[baseline=(X.base)]
\node[fill=green1, font=\bfseries, inner sep=5pt, rounded corners] (X) {0.601};\end{tikzpicture}                        \\
DeepSeek-R1               & \begin{tikzpicture}[baseline=(X.base)]
\node[fill=green3, inner sep=5pt, rounded corners] (X) {0.576};\end{tikzpicture}                        \\
Qwen3-30B-Instruct        & 0.425                        \\
Qwen3-30B-Thinking        & 0.485                        \\
Qwen3-235B-Instruct       & 0.554                        \\
Qwen3-235B-Thinking      & \begin{tikzpicture}[baseline=(X.base)]
\node[fill=green2, inner sep=5pt, rounded corners] (X) {0.586};\end{tikzpicture} \\ \bottomrule[1.5pt]                     
\end{tabular}
\caption{Dependency relationship identification performance (F1) of 13 LLMs. 
This task requires global reasoning over rule-to-rule dependencies (Sequential, Conditional, Parallel, None). 
}
\label{tab:accuracy results of dependency relationships}
\end{table}


For dependency reasoning (the right panel of Figure~\ref{fig:robustness_complexity}), performance divergence becomes more pronounced as the number of rule pairs grows.
Standard instruction-tuned models collapse under high reasoning complexity, whereas reasoning-oriented models exhibit significantly flatter degradation curves.
These results demonstrate that global rule flow modeling is not merely a pattern-matching task, but requires explicit state tracking and structured reasoning.

\begin{table*}[htbp]
    \setlength{\abovecaptionskip}{0.2cm}
    \setlength{\belowcaptionskip}{-0.3cm}
\centering
\small
\begin{tabular}{lccc}
\toprule
\textbf{Task} & \textbf{Qwen-30B (ExIde)} & \textbf{Qwen-30B (SFT)} & \textbf{$\Delta$} \\
\midrule
Rule Extraction & 83.90 & 84.70 & {\color{blue}+0.80} \\
Logic Operator Classifiction & 63.60 & 72.40 & {\color{blue}+8.80} \\
Dependency Relation Identification & 40.84 & 24.16 & {\color{red}-16.68} \\
\bottomrule
\end{tabular}
\caption{Performance comparison between the prompt-based ExIde framework and a fine-tuned baseline (SFT) on Qwen-30B.}
\label{tab:sft_results}
\end{table*}
For dependency reasoning (The right panel of Figure~\ref{fig:robustness_complexity}), performance divergence becomes more pronounced as the number of rule pairs grows.
Standard instruction-tuned models collapse under high reasoning complexity, whereas reasoning-oriented models exhibit significantly flatter degradation curves.
These results demonstrate that global rule flow modeling is not merely a pattern-matching task, but requires explicit state tracking and structured reasoning.

\subsection{Comparison with Supervised Fine-Tuning}
\label{sec:sft_comparison}

A natural question arises regarding the ceiling of our prompt-based pipeline compared to traditional model tuning. To investigate this, we conducted a Supervised Fine-Tuning (SFT) experiment using Low-Rank Adaptation (LoRA) \cite{DBLP:conf/iclr/HuSWALWWC22} on the Qwen-30B-Instruct model. We fine-tuned the model on the BREX dataset (300 samples for training and 109 samples for testing) to observe its performance dynamics across different sub-tasks.

As shown in Table \ref{tab:sft_results}, the results provide a fascinating insight into the trade-off between semantic extraction and structural reasoning, which we refer to as the \textit{Logic Gap}.

\paragraph{Semantic Extraction vs. Global Reasoning.} 
SFT indeed improves surface-level pattern recognition. It slightly enhances Rule Extraction (+0.8) and effectively learns domain-specific definitions for Logic Operators (+8.8). However, when tackling Dependency Relations, the SFT model suffers a severe degradation (-16.68) in global reasoning capabilities.

\paragraph{The ``Rule Pair Dropping'' Phenomenon.} 
Our error analysis reveals that this reasoning degradation is primarily driven by \textit{output incompleteness}. While ExIde consistently predicts relationships for all input pairs (e.g., given 10 pairs, it outputs 10 judgments), the SFT model frequently "gives up" halfway, truncating the output or missing pairs entirely. The Dependency Identification task requires global reasoning over a variable-length list of candidate pairs. SFT on BREX may caused the model to overfit to the average length or format of the training data, losing the generalization capability to handle the diverse complexity of the test set.

Ultimately, these findings validate the premise of our framework: while extracting atomic rules is fundamentally a semantic task where SFT is beneficial, constructing a dependency graph is a complex reasoning task. ExIde successfully leverages the innate, generalized reasoning power of Large Language Models, offering a far more robust solution for graph completeness than naive fine-tuning in low-resource, domain-specific settings.
\section{Conclusion}
In this work, we revisited business process modeling from a rule-centric perspective and identified a fundamental limitation of existing approaches: action-centric formulations fail to capture the complex logical dependencies that govern real-world business regulations.
To bridge this \emph{Logic Gap}, we introduced \textbf{BREX}, a cross-domain benchmark with expert-annotated condition--action rules and explicit rule-to-rule dependencies, enabling systematic evaluation of logic-aware extraction.
Building on this benchmark, we proposed \textbf{ExIde}, a structure-aware framework that decomposes business rule flow modeling into atomic rule extraction and global dependency reasoning.
Through extensive experiments across 13 large language models, we demonstrated that executable grounding provides a strong inductive bias for logic-intensive rule extraction, while reasoning-oriented models consistently outperform instruction-tuned counterparts in recovering long-range and non-linear dependencies.
Our findings suggest that effective modeling of complex business logic requires moving beyond surface-level semantic extraction toward execution-oriented representations that align with structured reasoning. We hope BREX will serve as a foundation for future research on logic-aware language models and advance the development of robust, automated process understanding systems.

\section{Limitations}

While our work makes progress toward logic-aware business rule flow modeling, several limitations remain.

\paragraph{Synthetic Data Usage.}
Although the majority of the BREX dataset is derived from real-world business documents, a limited amount of synthetic data is used to augment underrepresented logical structures such as deeply nested conditions and parallel constraints. Despite rigorous expert filtering and revision, synthetic texts may not fully capture the stylistic and contextual nuances of naturally occurring regulatory language. Future work could further reduce reliance on synthetic augmentation by expanding data collection in high-complexity domains.

\paragraph{Prompt-Based Dependency Modeling.}
ExIde relies on prompt-based reasoning to identify rule-to-rule dependencies rather than an end-to-end trained model. While this design choice enables flexible evaluation across diverse LLMs and avoids task-specific training, it may limit scalability when applied to extremely large rule sets due to the quadratic growth of rule pairs. Incorporating structural pruning or hybrid neural-symbolic approaches could improve efficiency in large-scale deployments.

\paragraph{Scope of Logical Formalization.}
Our annotation schema focuses on three primary dependency types (Sequential, Conditional, and Parallel) and a predefined set of logical operators. Although these categories cover the majority of business logic observed in our corpus, more expressive constructs—such as temporal constraints, exception handling, or probabilistic rules—are not explicitly modeled. Extending the schema to support richer logical formalisms remains an important direction for future research.

Despite these limitations, we believe our work represents an important step toward bridging natural language business rules and executable logical representations, and we hope it will stimulate further research on logic-aware reasoning with LLMs.

\section{Ethical Statement}

This work introduces BREX, a benchmark for business rule flow modeling, and proposes ExIde, a structure-aware reasoning framework for extracting structured business rules and their logical dependencies from unstructured regulatory and administrative texts. We discuss the ethical considerations associated with data collection, annotation, and potential downstream use.

\paragraph{Data Sources and Privacy.}
The BREX dataset is constructed primarily from publicly available or institutionally released business and regulatory documents, such as administrative guidelines, service manuals, and compliance descriptions. These documents do not contain personally identifiable information or sensitive user data. Any synthetic texts used for data augmentation are generated solely to enrich underrepresented logical structures and are carefully reviewed and revised by domain experts to ensure consistency with real-world regulatory language, without introducing fictitious entities or personal information.

\paragraph{Annotation Process and Human Involvement.}
All annotations in BREX are performed by trained domain experts following a clearly defined and documented annotation schema. Annotators are instructed to focus exclusively on extracting explicit business rules and logical dependencies present in the text, rather than inferring implicit intent or subjective interpretations. Inter-annotator agreement is evaluated to ensure consistency and reduce individual bias. No crowd-sourced or vulnerable populations are involved in the annotation process.

\paragraph{Potential Misuse and Limitations.}
While the proposed framework aims to support process automation and regulatory analysis, we acknowledge that automated extraction of business rules may be misused if deployed without appropriate human oversight, particularly in high-stakes domains such as finance, healthcare, or public administration. Incorrectly extracted or interpreted rules could lead to flawed downstream decisions. Therefore, we emphasize that BREX and ExIde are intended as research tools for benchmarking and model analysis, rather than as fully autonomous decision-making systems.

\paragraph{Model Behavior and Bias.}
Our study evaluates large language models using prompt-based methods without task-specific fine-tuning. As such, model outputs may reflect limitations or biases inherent in the underlying pretrained models. We do not claim that the extracted rule flows are universally correct or legally binding. Instead, our evaluation focuses on structural accuracy relative to expert-annotated ground truth, highlighting both strengths and failure modes to encourage transparent and responsible model development.

\paragraph{Broader Impact.}
We believe this work contributes positively to research on logic-aware language understanding and structured reasoning, by providing a realistic benchmark and systematic analysis of model behavior under increasing logical complexity. We hope that BREX will support future research on interpretable, verifiable, and human-in-the-loop systems, where automated rule extraction serves as an assistive tool rather than a replacement for expert judgment.
\section{Acknowledgement}
This research was partially supported by: Baima Lake Laboratory Joint Fund of the Zhejiang Provincial Natural Science Foundation of China (No.LBMHZ25F020001) and the National Natural Science Foundation of China (Grant No. 62276233).
\bibliography{custom}

\appendix
\section{Appendix}\label{sec:appendix}
\subsection{Definition}
\label{sec:definition}
\textbf{\textit{Definition 1}} A \textbf{Business Text} is a natural language description of a business scenario or process. It serves to standardize operations and management requirements within a specific domain.

\textbf{\textit{Definition 2}} \textbf{Business Rules} are explicit instructions describing the conditions, actions, and restrictions that must be followed. A business text typically contains multiple rules. We represent a rule as a binary tuple: $\langle$Condition, Action$\rangle$. Although complex rules may involve multiple conditions, we normalize them into atomic condition-action pairs linked through explicit dependency relations, enabling compositional modeling of complex logic. Beyond individual representation, capturing the \textbf{Dependency Relationships} between rules is crucial, as they dictate the logical execution flow (e.g., the execution of Rule B depends on the outcome of Rule A).

Based on these definitions, our annotation schema consists of three primary components:

\textbf{\textit{Definition 3}} \textbf{Condition} specifies the trigger for a business rule, comprising three key attributes: $\langle$Slot Type, Logical Operator, Reference Value$\rangle$.
\begin{itemize}[leftmargin=*, itemsep=0.1em, parsep=0pt, topsep=0.2em]
    \item \textbf{Slot Type}: Represents the category of user input (e.g., \textit{currency type}).
    \item \textbf{Logical Operator}: Defines the relation between the slot type and reference value. Permissible values include: \textit{contains}, \textit{equal to}, \textit{less than}, \textit{greater than}, \textit{less than or equal to}, and \textit{greater than or equal to}.
    \item \textbf{Reference Value}: Denotes specific data associated with the slot type, classified into:
    \begin{enumerate*}[label=(\alph*)]
        \item \textbf{Enumeration type} (discrete values, e.g., \textit{USD, RMB});
        \item \textbf{Numeric type} (values for comparison, e.g., \textit{\$50,000}).
    \end{enumerate*}
\end{itemize}

\textbf{\textit{Definition 4}} \textbf{Action} specifies the operation to be executed once the condition is met. If a process terminates, the Action value can be \textit{None}.

\textbf{\textit{Definition 5}} \textbf{Dependency Relationship} describes the logical connection between rules:
 \begin{itemize}[leftmargin=*, itemsep=0.1em, parsep=0pt, topsep=0.2em]
    \item \textbf{Sequential Dependency}: Rule A must be executed before Rule B.
    \item \textbf{Conditional Dependency}: Different reference values of Rule A trigger different subsequent rules (branching logic).
    \item \textbf{Parallel Dependency}: Two or more rules must be executed simultaneously.
 \end{itemize}

\subsection{Illustrative Examples of Dependency Relationships}
\label{sec:dependency relationship}

In this section, we provide illustrative examples of the three types of dependency relationships: sequential, conditional, and parallel. These examples aim to clarify the definitions discussed earlier, with corresponding business scenarios demonstrating the practical implications of each dependency type:
\begin{itemize}
    \item \textbf{Sequential Dependency:} Our bank supports up to 39 currency types of popular countries or regions around the world, including RMB, USD, JPY, GBP, and HKD. After selecting the appropriate currency, the customer needs to choose the corresponding cash or remittance type based on the type of business to be conducted thereafter. The cash or remittance type includes cash and remittance. Upon completing the cash/remittance type selection, the customer needs to provide the purchase amount...
    
    \item \textbf{Conditional Dependency:} When choosing a car rental service, the user first needs to select the vehicle type, which can be a sedan, SUV, business vehicle, or RV. If the user chooses a sedan or SUV, they will need to further choose the rental duration, with options like 1 day, 1 week, or 1 month. If the user chooses a business vehicle or RV, they will need to choose the rental purpose, which could be for travel, road trips, or business activities. After selecting the rental duration, the user will also need to choose whether to purchase insurance...
    
    \item \textbf{Parallel Dependency:} When signing up for an educational training course, the user first needs to choose the course type, such as programming, design, marketing, etc. After selecting the course type, the user needs to simultaneously choose the instruction mode and course duration. The instruction mode includes online courses and in-person courses, and the course duration can be 1 month, 3 months, or 6 months. After selecting the instruction mode and course duration, the user will need to choose the learner's age group, which could be adults or teenagers. Then, the user will proceed to the tuition payment...
\end{itemize} 

\begin{figure}[t]
    \centering
    \includegraphics[width=0.25\textwidth]{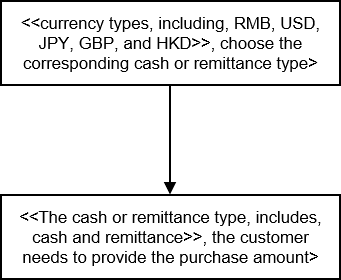}
    \caption{Sequential dependency.}
    \label{fig:sequential dependency}
\end{figure}

\begin{figure}[t]
    \centering
    \includegraphics[width=0.45\textwidth]{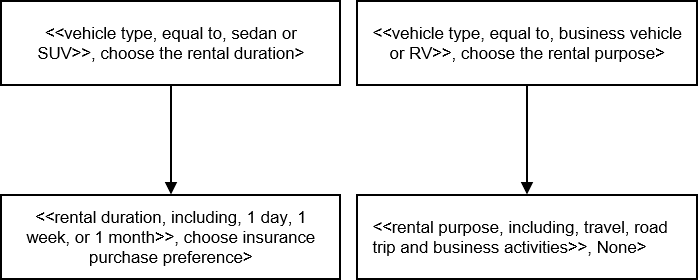}
    \caption{Conditional dependency.}
    \label{fig:conditional dependency}
\end{figure}

\begin{figure}[t]
    \centering
    \includegraphics[width=0.45\textwidth]{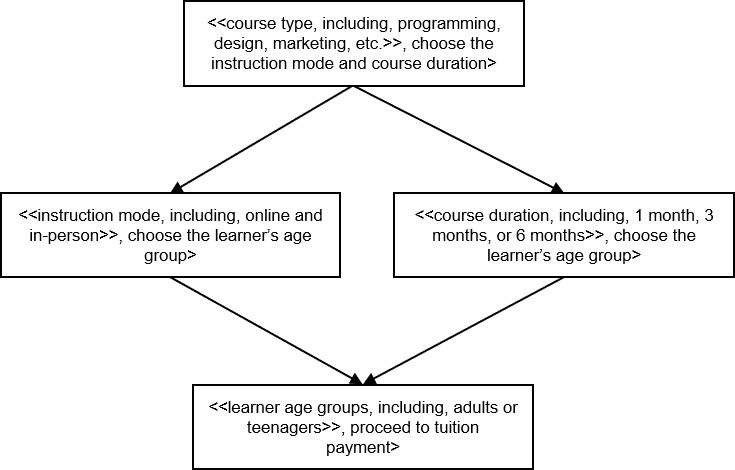}
    \caption{Parallel dependency.}
    \label{fig:parallel dependency}
\end{figure}
Figures \ref{fig:sequential dependency}, \ref{fig:conditional dependency}, and \ref{fig:parallel dependency} visually represent the business processes associated with each type of dependency.

A sequential dependency occurs when the execution of one rule depends on the completion of another rule. In other words, the first rule must be executed before the second. For example, in the case of foreign exchange services, a customer must first select a currency type (such as RMB, USD, JPY, etc.). After selecting the currency, the customer must choose the corresponding cash or remittance type (cash or remittance). Finally, after selecting the cash/remittance type, the customer is required to provide the exchange amount. Each step must occur in a prescribed order, as the actions are dependent on prior selections, as depicted in Fig. \ref{fig:sequential dependency}.

A conditional dependency is present when the next rule to be executed depends on the conditions or choices made in the previous rule. In this scenario, the execution of a rule is conditional on the reference values selected by the user. For instance, in the car rental service example, when a user selects a vehicle type, the subsequent actions depend on the choice made. If the user selects a sedan or SUV, they are prompted to choose the rental duration, with options like 1 day, 1 week, or 1 month. However, if the user selects a business vehicle or RV, they will instead need to choose the rental purpose (e.g., travel, road trip, or business activity). This dependency ensures that the process flow adjusts based on user input, as shown in Fig. \ref{fig:conditional dependency}.

A parallel dependency exists when multiple business rules are executed simultaneously. For example, when registering for an educational training course, a user first selects a course type (such as programming, design, marketing, etc.). After this selection, the user must choose both the instruction mode (online or in-person) and the course duration (1 month, 3 months, or 6 months) in parallel. Then, the learner's age group (either adults or teenagers) is selected. Finally, the user proceeds to tuition payment after completing these selections. These choices of instruction mode and course duration occur in parallel, as illustrated in Fig. \ref{fig:parallel dependency}.

\begin{table*}[t]
    \renewcommand{\arraystretch}{1.1}
    \setlength{\tabcolsep}{4pt} 
    \centering
    \small
    \begin{tabular}{c|c|c|c|c|c|c|c|c|c|c|c|c} \toprule[1.2pt]
    \multirow{2}{*}{\textbf{Docs}} & \multirow{2}{*}{\textbf{Sentences}} & \multirow{2}{*}{\textbf{Rules}} & \multirow{2}{*}{\textbf{Tokens}} & \multicolumn{6}{c|}{\textbf{Logical Operator}} & \multicolumn{3}{c}{\textbf{Dependency Relationship}} \\ \cmidrule(lr){5-10} \cmidrule(lr){11-13}
     & & & & Contains & Equal & < & > & $\le$ & $\ge$ & Sequential & Conditional & Parallel \\ \midrule
    409 & 2,573 & 2,855 & 95,779 & 1,798 & 41 & 34 & 900 & 38 & 44 & 2,330 & 722 & 417 \\ \bottomrule[1.2pt]     
    \end{tabular}
    \caption{Statistics of the BREX dataset. We present statistical information on the number of documents and various elements in the dataset.}
    \label{tab:statistical information}
\end{table*}
\subsection{Detailed Dataset Statistics}
\label{sec:detailed_statistics}
To provide a granular view of the BREX benchmark, we present detailed statistics in Table \ref{tab:statistical information}. The dataset comprises 409 documents with a total of 95,779 tokens and 2,573 sentences, averaging approximately 6.3 sentences and 234 tokens per document. This reflects the concise and information-dense nature of professional regulatory texts.

In terms of \textbf{Logical Operators}, the distribution reveals the complexity of real-world constraints. Notably, \emph{Contains} (1,798) and \emph{Greater Than} ($>$) (900) are the most predominant operators, accounting for the majority of conditions. This indicates that business rules frequently involve set membership checks (e.g., verifying if a product belongs to a specific category) and numerical threshold constraints (e.g., financial limits), rather than simple equality matches.

Regarding \textbf{Dependency Relationships}, while \emph{Sequential} dependencies (2,330) form the backbone of most processes, non-linear structures constitute a significant portion of the logic flow. Specifically, \emph{Conditional} (722) and \emph{Parallel} (417) dependencies together account for approximately \textbf{33\%} of all relations. This high proportion of branching and concurrent logic empirically supports our motivation that modeling business regulations requires capabilities far beyond linear action sequencing.

\subsection{Benchmark Comparison}
\label{sec:benchmark_comparison}

To further clarify the positioning of BREX within the broader landscape of procedural text understanding, we provide a systematic comparison with existing benchmarks in Table \ref{tab:benchmark_comparison}.

Existing datasets predominantly follow an \emph{Action-Centric} paradigm, treating processes as sequences of events or actions (e.g., recipes, maintenance logs). While some recent works like PAGED \cite{DBLP:conf/acl/DuLLL24} have scaled up the data size significantly, they rely on synthetic data-to-text generation, which often simplifies the linguistic complexity found in real-world regulations. In contrast, BREX is the first benchmark to adopt a \emph{Rule-Centric} modeling approach, explicitly annotating atomic business rules and their complex logical dependencies (Sequential, Conditional, Parallel) across 30+ real-world vertical domains.

\subsection{Text Quality Assessment Criteria}
\label{sec:text_quality_assessment_criteria}
\begin{figure}[t]
    \centering
    \includegraphics[width=0.45\textwidth]{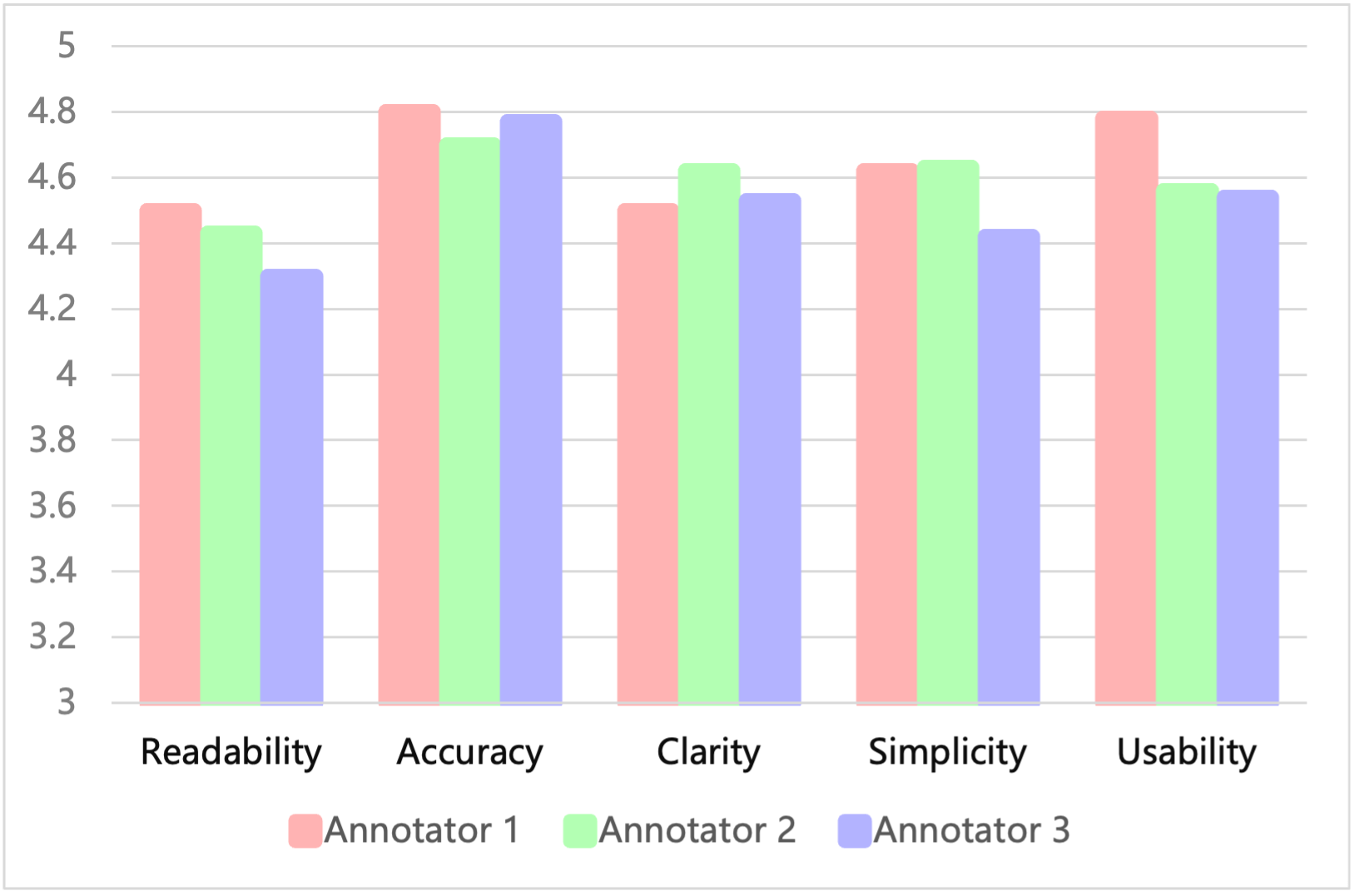}
    \caption{Human evaluation results of text quality across five dimensions among three annotators.}
    \label{ICC}
\end{figure}

To assess the linguistic and practical quality of the business texts in BREX, we conduct a human evaluation along five dimensions: \textit{Readability}, \textit{Accuracy}, \textit{Clarity}, \textit{Simplicity}, and \textit{Usability}. 
Each dimension is rated independently from 1 to 5 by trained annotators. 
Below we provide detailed definitions of each dimension, followed by the scoring guidelines used during annotation to ensure consistency and reproducibility. Evaluation results can be seen in Figure \ref{ICC}.

\subsection{Details on Inter-Annotator Agreement Calculation}
\label{sec:iaa_details}

To quantitatively evaluate the reliability of BREX, we addressed the challenge that standard metrics for inter-annotator agreement (IAA) are not directly applicable to complex, graph-structured dependency data. Consequently, we adopted a proxy evaluation strategy by projecting the structured annotations into a flattened Named Entity Recognition (NER) format.

As illustrated in Table \ref{tab:ner_annotation_example}, we decomposed the hierarchical rule structures into token-level span annotations using the standard BIO (Beginning, Inside, Outside) tagging scheme. We focused on the three atomic components of business rules: \textbf{Slot Types}, \textbf{Reference Values}, and \textbf{Actions}. 

The table demonstrates this projection with a representative example. For the phrase \textit{``currency types... including RMB, USD...''}:
\begin{itemize}
    \item \textbf{Span Boundaries}: Annotators must determine the exact boundaries of slots (e.g., ``currency'' vs. ``currency types''). In the example, Annotator 2 marks ``types'' as \texttt{O} (Outside), while Annotators 1 and 3 include it as \texttt{I-Slot}.
    \item \textbf{Entity Separation}: Annotators must also distinguish between distinct values. For the enumeration ``RMB, USD'', Annotator 2 identifies them as separate entities (\texttt{B-Ref}, \texttt{B-Ref}), representing a fine-grained interpretation, whereas Annotators 1 and 3 might treat the sequence differently.
\end{itemize}

This linearization transforms the structural agreement problem into a token-level classification problem, enabling the calculation of Fleiss' Kappa \cite{artstein2017inter}. The resulting high agreement score (0.901) confirms that, despite minor boundary variations (as seen in the table), the core semantic definitions of business rules are interpreted consistently across experts.

\paragraph{Evaluation Dimensions.}
\begin{itemize}[leftmargin=*]
    \item \textit{Readability}: The extent to which a business text can be easily read and understood by a general user, considering sentence structure, vocabulary choice, and overall linguistic fluency.
    
    \item \textit{Accuracy}: The degree to which the business text correctly and faithfully describes the underlying business service, rules, and constraints, without introducing factual errors, omissions, or misleading information.
    
    \item \textit{Clarity}: The extent to which the text presents business rules and execution logic in a clear and logically coherent manner, enabling readers to understand conditional branching, rule dependencies, and overall process flow.
    
    \item \textit{Simplicity}: The degree to which the text avoids unnecessary redundancy, excessive verbosity, or irrelevant details, while preserving all essential information required to describe the business process.
    
    \item \textit{Usability}: The effectiveness of the text in guiding users to successfully complete the intended business service or procedure, from an operational and instructional perspective.
\end{itemize}

All annotators are instructed to apply these criteria consistently across documents and dimensions. 
Inter-annotator agreement is reported using the Intraclass Correlation Coefficient (ICC) in the main paper, demonstrating high annotation reliability.

\begin{table*}[t]
\resizebox{\textwidth}{!}{
\begin{tabular}{lrcl}
\toprule
\textbf{Dataset}                                              & \textbf{Samples}                                                              & \textbf{Paradigm}           & \textbf{Source}                                          \\ \midrule
\citet{quishpi2020extracting}       & 121                                                                  & Action-Flow        & Tutorials, maintenance manuals                     \\
\citet{qian2020approach}            & 360                                                                  & Action-Flow        & Cooking recipes and maintenance manuals            \\
\citet{ackermann2021data}           & 358                                                                  & Action-Flow        & Cooking recipes and maintenance   manuals  \\
\citet{lopez2021declarative}        & 37                                                                   & Action-Flow        &   4 entries from the BPM Academic Initiative                                            \\
\citet{DBLP:conf/aiia/BellanGDPA22} & 45                                                                   & Action-Flow        & Academic, Industry, Textbook, Public Sector        \\
\citet{liang-etal-2023-knowing}     & 200                                                                  & Action-Flow        & User manuals, e-commerce support documents \\
\citet{ren2023constructing}         & 283                                                                  & Action-Flow        & Online wikiHow knowledge base                      \\
\citet{DBLP:conf/acl/DuLLL24}                 & 3,394                                                                 & Action-Flow        & \textbf{Synthetic} (Data-to-Text) \\ \midrule
\textbf{BREX (Ours)}                                          & 409 & Rule-Flow          & \textbf{Real-world} (+Aug)                                         \\
& (2,855 rules)    & (Condition-Action) & (30+Domains) \\ \bottomrule 
\end{tabular}}
\caption{Comparison of BREX with existing procedural text and process extraction benchmarks. \textbf{Paradigm} distinguishes between Action-Flow (extracting event sequences) and Rule-Flow (extracting logical constraints). \textbf{Source} indicates whether the text is naturally occurring (Real) or generated (Synthetic).}
\label{tab:benchmark_comparison}
\end{table*}

\begin{table*}[t]
    \renewcommand{\arraystretch}{1.1}
    \centering
    \small
    
    \scalebox{1}{
        \begin{tabular}{c|c|c|c|c|c|c|c|c|c|c|c} \toprule[1.2pt]
          Text Tokens & \dots & currency & types & of & \dots & popular & including & RMB & USD & JPY & \dots \\ \midrule
          Annotator 1 & \dots & B-Slot & I-Slot & O & \dots & O & O & B-Ref & I-Ref & I-Ref & \dots \\ 
          Annotator 2 & \dots & B-Slot & O & O & \dots & O & O & B-Ref & B-Ref & B-Ref & \dots \\ 
          Annotator 3 & \dots & B-Slot & I-Slot & O & \dots & O & O & B-Ref & I-Ref & I-Ref & \dots \\ \bottomrule[1.2pt] 
        \end{tabular}} 
        \caption{An example of transforming rule annotation into NER format for Inter-Annotator Agreement (IAA) calculation.}
        \label{tab:ner_annotation_example}
\end{table*}

\subsection{Case Study}
\label{sec:case_study}
\begin{table*}[htbp]
    \centering
    \renewcommand{\arraystretch}{1.3}
    \begin{tabularx}{\textwidth}{@{}lX@{}}
        \toprule
        \textbf{Category} & \textbf{Content} \\
        \midrule
        \textbf{Input Text} & When employees apply for leave in the company, they first need to clarify the type of leave, which mainly includes personal leave, sick leave and annual leave. After choosing the type of leave, employees need to submit the duration of their leave. \\
        \midrule
        
        \textbf{Ground Truth} \newline  \twemoji{glowing star}  & 
        1. $\langle\langle$Type of leave, include, personal leave, sick leave and annual leave$\rangle$, None$\rangle$ \newline
        2. $\langle\langle$Type of leave, equal, personal leave$\rangle$, employees need to submit the duration of their leave$\rangle$ \newline
        3. $\langle\langle$Type of leave, equal, sick leave$\rangle$, employees need to submit the duration of their leave$\rangle$ \newline
        4. $\langle\langle$Type of leave, equal, annual leave$\rangle$, employees need to submit the duration of their leave$\rangle$ \\
        \midrule
        \textbf{Good Output} \newline \twemoji{check mark button} & 
        1. $\langle\langle$Type of leave, include, personal leave, sick leave and annual leave$\rangle$, None$\rangle$ \newline
        2. $\langle\langle$Type of leave, equal, personal leave$\rangle$, employees need to submit the duration of their leave$\rangle$ \newline
        3. $\langle\langle$Type of leave, equal, sick leave$\rangle$, employees need to submit the duration of their leave$\rangle$ \newline
        4. $\langle\langle$Type of leave, equal, annual leave$\rangle$, employees need to submit the duration of their leave$\rangle$ \\
        \midrule
        \textbf{Bad Output} \newline \twemoji{cross mark} & 
        1. $\langle\langle$Type of leave, include, personal leave, sick leave and annual leave$\rangle$, None$\rangle$ \newline
        2. $\langle\langle$Type of leave, equal, personal leave, sick leave and annual leave$\rangle$, employees need to submit the duration of their leave$\rangle$ \\
        \bottomrule
    \end{tabularx}
    \caption{Case~1: Over-merging enumerated values under the \textit{equal} operator.
Distinct leave types with identical downstream actions are incorrectly collapsed
into a single equality condition, violating the atomic semantics of equality and
reducing rule granularity.}
    \label{tab:leave_logic}
\end{table*}

\paragraph{Case 1 \& Case 2: Over-Merging Enumerated Values under the \textit{equal} Operator.}

The failures illustrated in Table~\ref{tab:leave_logic} and Table~\ref{tab:credit_card_logic} are caused by an \textbf{over-aggressive merging of enumerated values under the \textit{equal}
operator}. In Table~\ref{tab:leave_logic}, the business text specifies that after selecting \emph{any} leave type—personal leave, sick leave, or annual leave—the employee must submit the duration of the leave. Although the downstream action is identical, each leave type corresponds to a distinct equality condition and is annotated as a separate rule in the ground truth. In Table~\ref{tab:time_deposit_logic}, the business text explicitly specifies two separate decision branches: after selecting a \emph{standard credit card} or a \emph{themed co-branded card}, the applicant proceeds to the same downstream action. Despite the identical action, the two branches correspond to distinct conditional instantiations and are annotated as separate rules in the ground truth.

Under Prompt 4, some models incorrectly collapse all enumerated leave types into a single compound equality condition (e.g., \textit{equal(personal, sick, annual)}). This violates the intended semantics of the rule schema, where the \textit{equal} operator is defined over a single atomic value rather than a set. As a result, multiple rule instantiations are replaced by a single abstract rule, leading to a loss of rule granularity. While the generated output appears more concise, it fails to preserve the one-to-one correspondence between decision options and rule instances required by the annotation standard, resulting in reduced recall and lower F1 scores.

\begin{table*}[htbp]
    \centering
    \renewcommand{\arraystretch}{1.3}
    \begin{tabularx}{\textwidth}{@{}lX@{}}
        \toprule
        \textbf{Category} & \textbf{Content} \\
        \midrule
        \textbf{Input Text} & When applying for our bank's credit card, the applicant needs to first select the type of card they wish to apply for. Our bank offers two options: standard credit cards and themed co-branded cards. After choosing the standard credit card, the applicant needs to fill in their personal basic information. \dots After choosing the theme co-branded card, the applicant also needs to fill in their personal basic information. \\
        \midrule
        \textbf{Ground Truth} \newline \twemoji{glowing star} & 
        1. $\langle\langle$Type of card, include, standard credit cards and themed co-branded cards$\rangle$, None$\rangle$ \newline
        2. $\langle\langle$Type of card, equal, standard credit card$\rangle$, the applicant needs to fill in their personal basic information$\rangle$ \newline
        3. $\langle\langle$Type of card, equal, theme co-branded card$\rangle$, the applicant also needs to fill in their personal basic information$\rangle$ \\
        \midrule
        \textbf{Good Output} \newline \twemoji{check mark button} & 
        1. $\langle\langle$Type of card, include, standard credit cards and themed co-branded cards$\rangle$, None$\rangle$ \newline
        2. $\langle\langle$Type of card, equal, standard credit card$\rangle$, the applicant needs to fill in their personal basic information$\rangle$ \newline
        3. $\langle\langle$Type of card, equal, theme co-branded card$\rangle$, the applicant also needs to fill in their personal basic information$\rangle$ \\
        \midrule
        \textbf{Bad Output} \newline \twemoji{cross mark} & 
        1. $\langle\langle$Type of card, include, standard credit cards and themed co-branded cards$\rangle$, None$\rangle$ \newline
        2. $\langle\langle$Type of card, equal, standard credit card, theme co-branded card$\rangle$, the applicant needs to fill in their personal basic information$\rangle$ \\
        \bottomrule
    \end{tabularx}
    \caption{Case~2: Over-merging enumerated values under the \textit{equal} operator.
Distinct card types with identical downstream actions are incorrectly collapsed
into a single equality condition, violating the atomic semantics of equality and
reducing rule granularity.}
    \label{tab:credit_card_logic}
\end{table*}




\begin{table*}[t]
    \centering
    \renewcommand{\arraystretch}{1.3}
    \begin{tabularx}{\textwidth}{@{}lX@{}}
        \toprule
        \textbf{Category} & \textbf{Content} \\
        \midrule
        \textbf{Input Text} & When handling a time deposit at a bank, the customer must first choose the \begin{tikzpicture}[baseline=(X.base)]
\node[fill=red!20, inner sep=2pt, rounded corners] (X) {type of deposit}; \end{tikzpicture}, which mainly includes two options: lump-sum deposit with lump-sum withdrawal and installment deposit with lump-sum withdrawal. If the customer chooses a lump-sum deposit with lump-sum withdrawal, they must first determine the deposit currency. The bank supports currencies including the Chinese yuan (RMB), U.S. dollars, and euros. Next, the customer selects the deposit term, which can be one year, three years, or five years. If the customer chooses an installment deposit with lump-sum withdrawal, they must first set a fixed \begin{tikzpicture}[baseline=(X.base)]
\node[fill=green!20, inner sep=2pt, rounded corners] (X) {monthly deposit amount }; \end{tikzpicture}, which must be no less than 50 yuan. After that, the customer selects the deposit term, which can be either one year or three years. Once the \begin{tikzpicture}[baseline=(X.base)]
\node[fill=orange!20, inner sep=2pt, rounded corners] (X) {above information}; \end{tikzpicture} is confirmed, the customer needs to provide the specific deposit amount. If a single deposit exceeds 200,000 yuan, the customer must additionally provide proof of the source of funds; if the amount is less than or equal to 200,000 yuan, no such proof is required. Finally, the customer may choose whether to activate \begin{tikzpicture}[baseline=(X.base)]
\node[fill=cyan!20, inner sep=2pt, rounded corners] (X) {the automatic renewal service and the method for receiving interest rate}; \end{tikzpicture} \begin{tikzpicture}[baseline=(X.base)]
\node[fill=cyan!20, inner sep=2pt, rounded corners] (X) {change notifications }; \end{tikzpicture}. The \begin{tikzpicture}[baseline=(X.base)]
\node[fill=yellow!20, inner sep=2pt, rounded corners] (X) {automatic renewal service }; \end{tikzpicture} can be set to ``Yes'' or ``No,'' while \begin{tikzpicture}[baseline=(X.base)]
\node[fill=blue!20, inner sep=2pt, rounded corners] (X) {notification methods}; \end{tikzpicture} include ``SMS notification'' and ``email notification.'' These two services can be selected simultaneously. \\
        \midrule
        \textbf{Ground Truth} \newline \twemoji{glowing star} & 
        1. $\langle\langle$ \begin{tikzpicture}[baseline=(X.base)]
\node[fill=red!20, inner sep=2pt, rounded corners] (X) {Type of deposit}; \end{tikzpicture}, include, lump-sum deposit with lump-sum withdrawal and installment deposit with lump-sum withdrawal$\rangle$, None$\rangle$ \newline
        2. $\langle\langle$ \begin{tikzpicture}[baseline=(X.base)]
\node[fill=green!20, inner sep=2pt, rounded corners] (X) {Monthly deposit amount}; \end{tikzpicture}, $>$, 200,000$\rangle$, the customer must additionally provide proof of the source of funds$\rangle$ \newline
        3. $\langle\langle$ \begin{tikzpicture}[baseline=(X.base)]
\node[fill=yellow!20, inner sep=2pt, rounded corners] (X) {Automatic renewal service}; \end{tikzpicture}, include, ``Yes'' or ``No,''$\rangle$, None$\rangle$ \newline
        4. $\langle\langle$ \begin{tikzpicture}[baseline=(X.base)]
\node[fill=blue!20, inner sep=2pt, rounded corners] (X) {Notification methods}; \end{tikzpicture}, include, ``SMS notification'' and ``email notification.''$\rangle$, None$\rangle$ \newline
        ...\\
        \midrule
        \textbf{Prompt 4 - Good Output} \newline \twemoji{check mark button} & 
        1. $\langle\langle$ \begin{tikzpicture}[baseline=(X.base)]
\node[fill=red!20, inner sep=2pt, rounded corners] (X) {Type of deposit}; \end{tikzpicture}, include, lump-sum deposit with lump-sum withdrawal and installment deposit with lump-sum withdrawal$\rangle$, None$\rangle$ \newline
        2. $\langle\langle$ \begin{tikzpicture}[baseline=(X.base)]
\node[fill=green!20, inner sep=2pt, rounded corners] (X) {Monthly deposit amount}; \end{tikzpicture}, $>$, 200,000$\rangle$, the customer must additionally provide proof of the source of funds$\rangle$ \newline
        3. $\langle\langle$ \begin{tikzpicture}[baseline=(X.base)]
\node[fill=yellow!20, inner sep=2pt, rounded corners] (X) {Automatic renewal service}; \end{tikzpicture}, include, ``Yes'' or ``No,''$\rangle$, None$\rangle$ \newline
        4. $\langle\langle$ \begin{tikzpicture}[baseline=(X.base)]
\node[fill=blue!20, inner sep=2pt, rounded corners] (X) {Notification methods}; \end{tikzpicture}, include, ``SMS notification'' and ``email notification.''$\rangle$, None$\rangle$ \newline
        ...\\
        \midrule
        \textbf{Prompt 5 - Bad Output} \newline \twemoji{cross mark} & 
        1. $\langle\langle$ \begin{tikzpicture}[baseline=(X.base)]
\node[fill=orange!20, inner sep=2pt, rounded corners] (X) {Above information }; \end{tikzpicture}, include, type of deposit, the deposit currency, deposit term and monthly deposit amount$\rangle$, customer needs to provide the specific deposit amount$\rangle$ \newline
        2. $\langle\langle$ \begin{tikzpicture}[baseline=(X.base)]
\node[fill=cyan!20, inner sep=2pt, rounded corners] (X) {The automatic renewal service and the method for receiving interest}; \end{tikzpicture} \begin{tikzpicture}[baseline=(X.base)]
\node[fill=cyan!20, inner sep=2pt, rounded corners] (X) {rate change notifications }; \end{tikzpicture}, include, automatic renewal service, notification methods$\rangle$, These two services can be selected simultaneously.$\rangle$ \newline
        ...\\
        \bottomrule
    \end{tabularx}
    \caption{Case~3: Abstraction-induced precision degradation in high-complexity texts.
Prompt 5 reifies discourse-level connectors into abstract slots and meta-rules,
leading to spurious extractions, while Prompt 4 remains grounded in text-aligned
fields and achieves higher precision.}
    \label{tab:time_deposit_logic}
\end{table*}

\paragraph{Case 3: Abstraction-Induced Precision Degradation in High-Complexity Texts.}
Table~\ref{tab:time_deposit_logic} presents a representative High-complexity example involving a dense fixed-deposit process with multiple branching conditions, numeric thresholds, and auxiliary services, resulting in a high rule count (15 rules). In this setting, Prompt 4 consistently grounds its outputs in
\textbf{explicit, surface-level slots and thresholds directly stated in the text}, such as deposit type, deposit amount constraints, automatic renewal options, and notification methods. By discouraging abstraction beyond textually grounded entities, Prompt 4 produces rule sets that closely align with the
annotated schema and maintain high precision.

In contrast, Prompt 5's pseudo-code paradigm introduces a distinct failure mode under high logical density. By enforcing a procedural abstraction, Prompt 5 encourages the model to \textbf{reify discourse-level connective phrases}—such as ``once the above information is confirmed'' or ``these two services can be selected simultaneously''—into variables or slot-like entities. This results in
the generation of abstract or meta-level rules (e.g., slots like \emph{``above information''} or composite service entities) that do not exist in the ground truth annotation. While such rules may appear structurally coherent within the pseudo-code framework, they constitute spurious extractions under
tuple-level evaluation and directly reduce precision.

This phenomenon is amplified in high-density texts, where narrative connectors and summarizing statements are more frequent. As a result, Prompt 5 systematically over-generates abstract slots and meta-rules, leading to a pronounced precision drop, particularly for smaller or less robust models.
Prompt 4, by contrast, retains a more conservative extraction bias, favoring concrete, text-aligned rules over procedural abstraction. This conservative bias enables Prompt 4 to outperform Prompt 5 in the High bucket, despite its higher variance in lower-complexity regimes.
\clearpage
\begin{figure*}[t]
    \centering
    \includegraphics[width=1\textwidth]{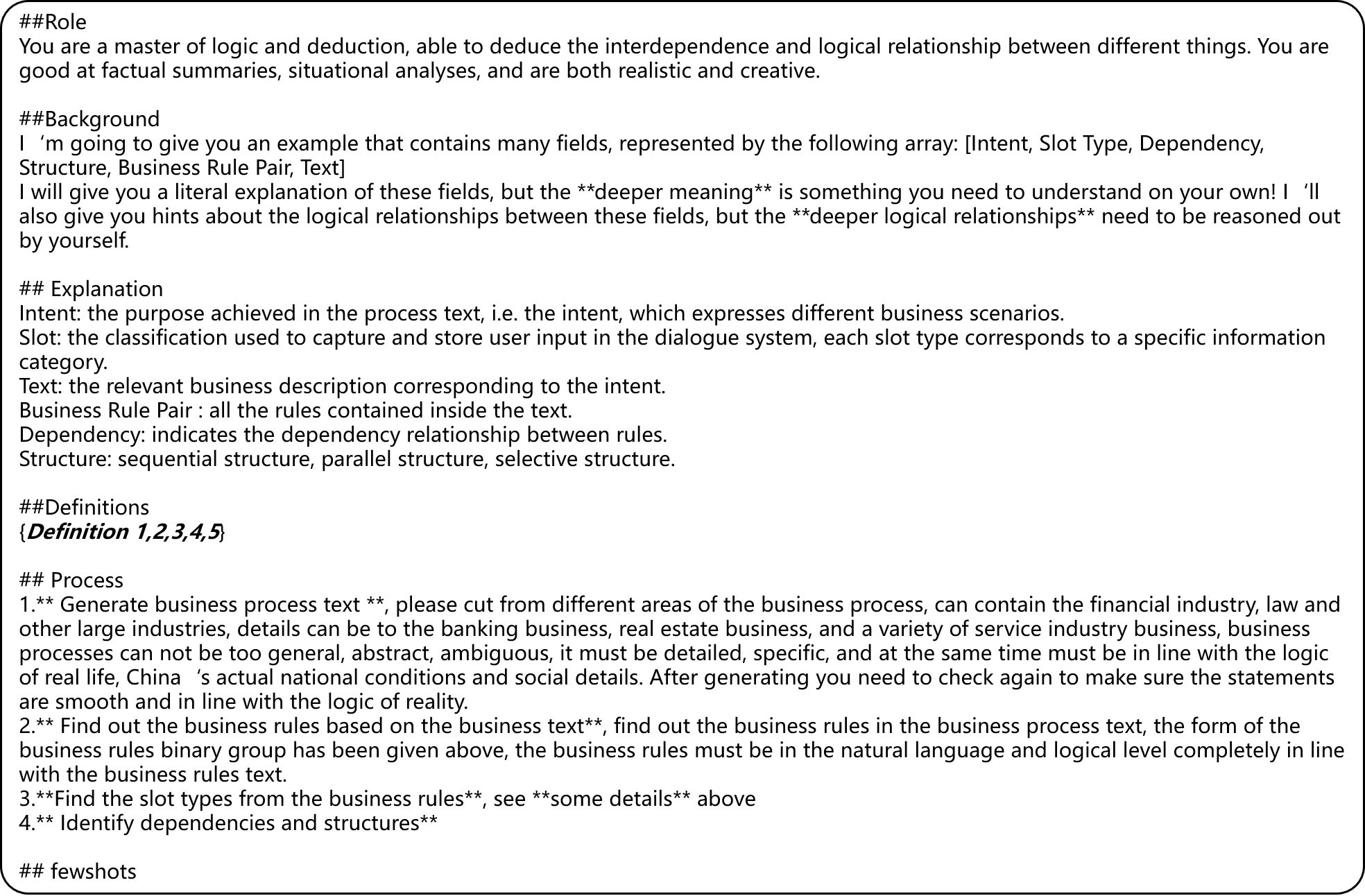}
    \caption{The prompt of synthetic text generation.}
    \label{fig:data generation prompt}
\end{figure*}

\begin{figure*}[t]
    \centering
    \includegraphics[width=1\textwidth]{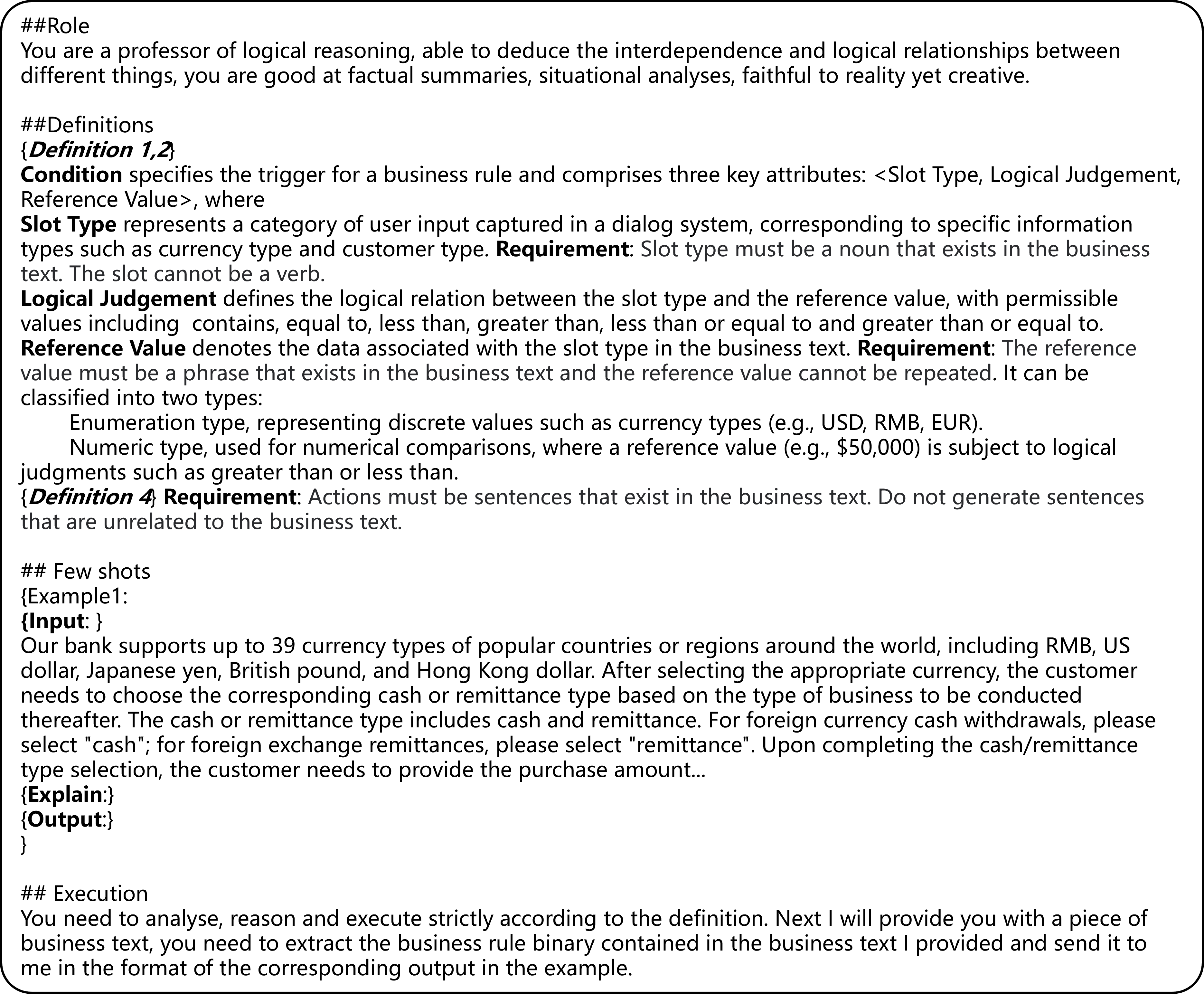}
    \caption{The basic template of five prompts.}
    \label{fig:5 prompt template}
\end{figure*}

\begin{figure*}[t]
    \centering
    \includegraphics[width=1\textwidth]{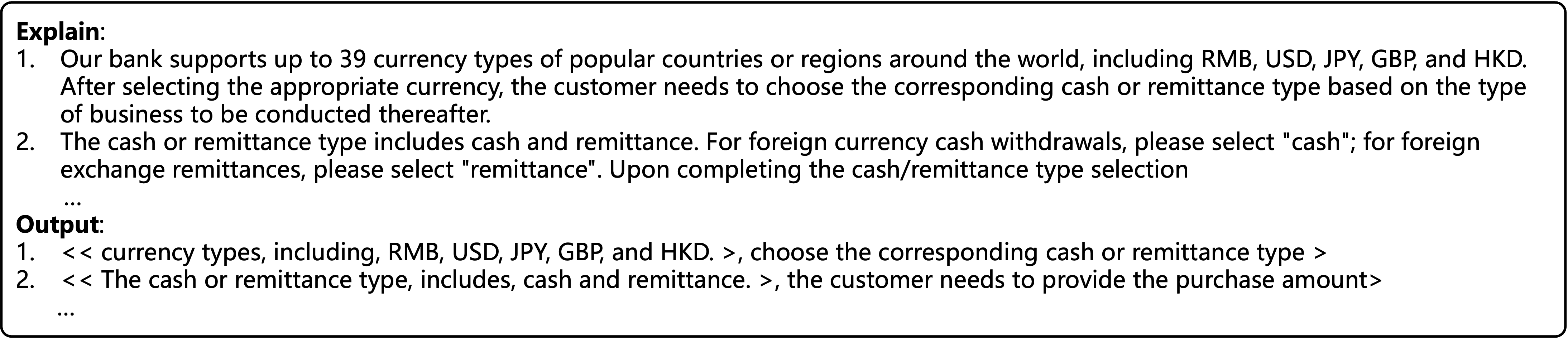}
    \caption{The details for Prompt 1 (Implicit Semantic Alignment).}
    \label{fig:prompt1}
\end{figure*}

\begin{figure*}[t]
    \centering
    \includegraphics[width=1\textwidth]{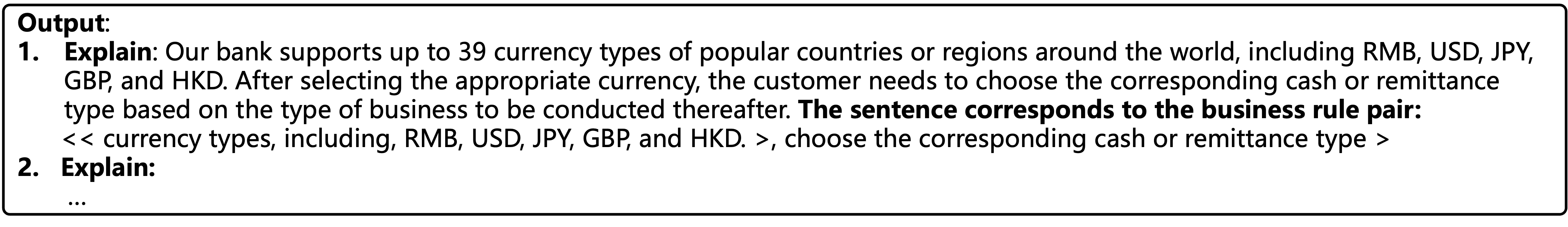}
    \caption{The details for Prompt 2 (Explicit Traceability).}
    \label{fig:prompt2}
\end{figure*}

\begin{figure*}[t]
    \centering
    \includegraphics[width=1\textwidth]{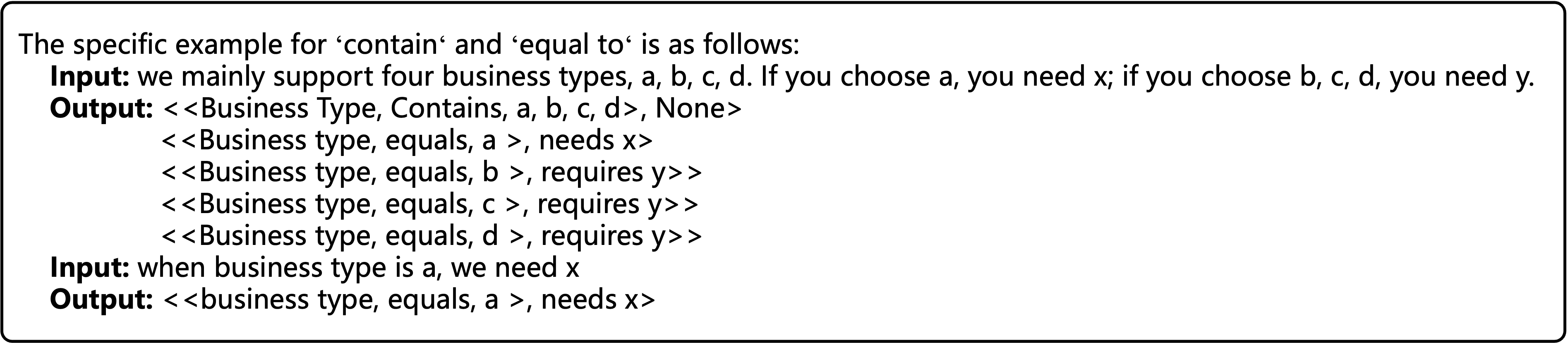}
    \caption{The details for Prompt 4 (Logic-Aware Definition Injection).}
    \label{fig:prompt3}
\end{figure*}

\begin{figure*}[t]
    \centering
    \includegraphics[width=1\textwidth]{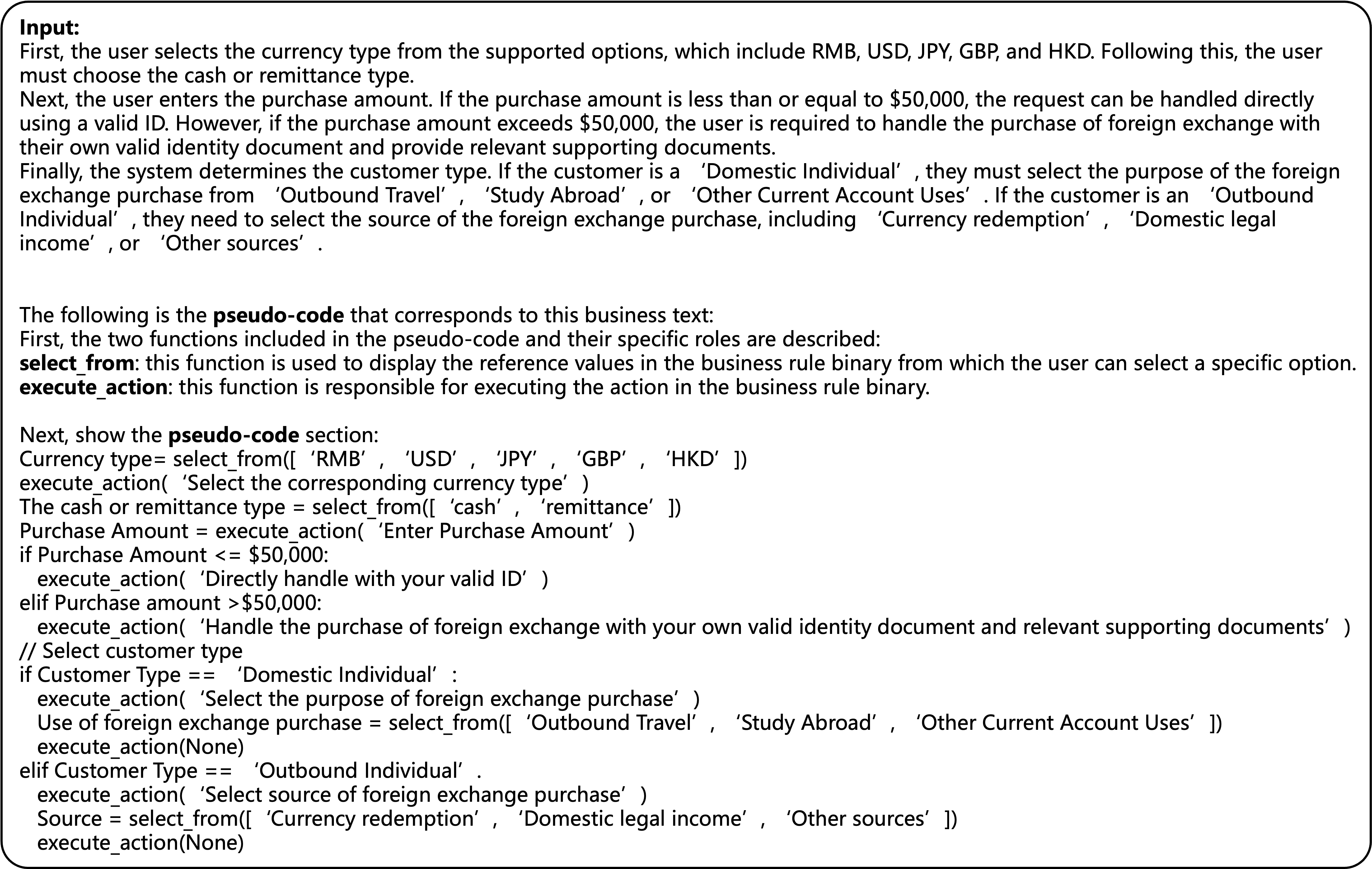}
    \caption{The details for Prompt 5 (Executable Grounding).}
    \label{fig:prompt5}
\end{figure*}

\begin{figure*}[t]
    \centering
    \includegraphics[width=1\textwidth]{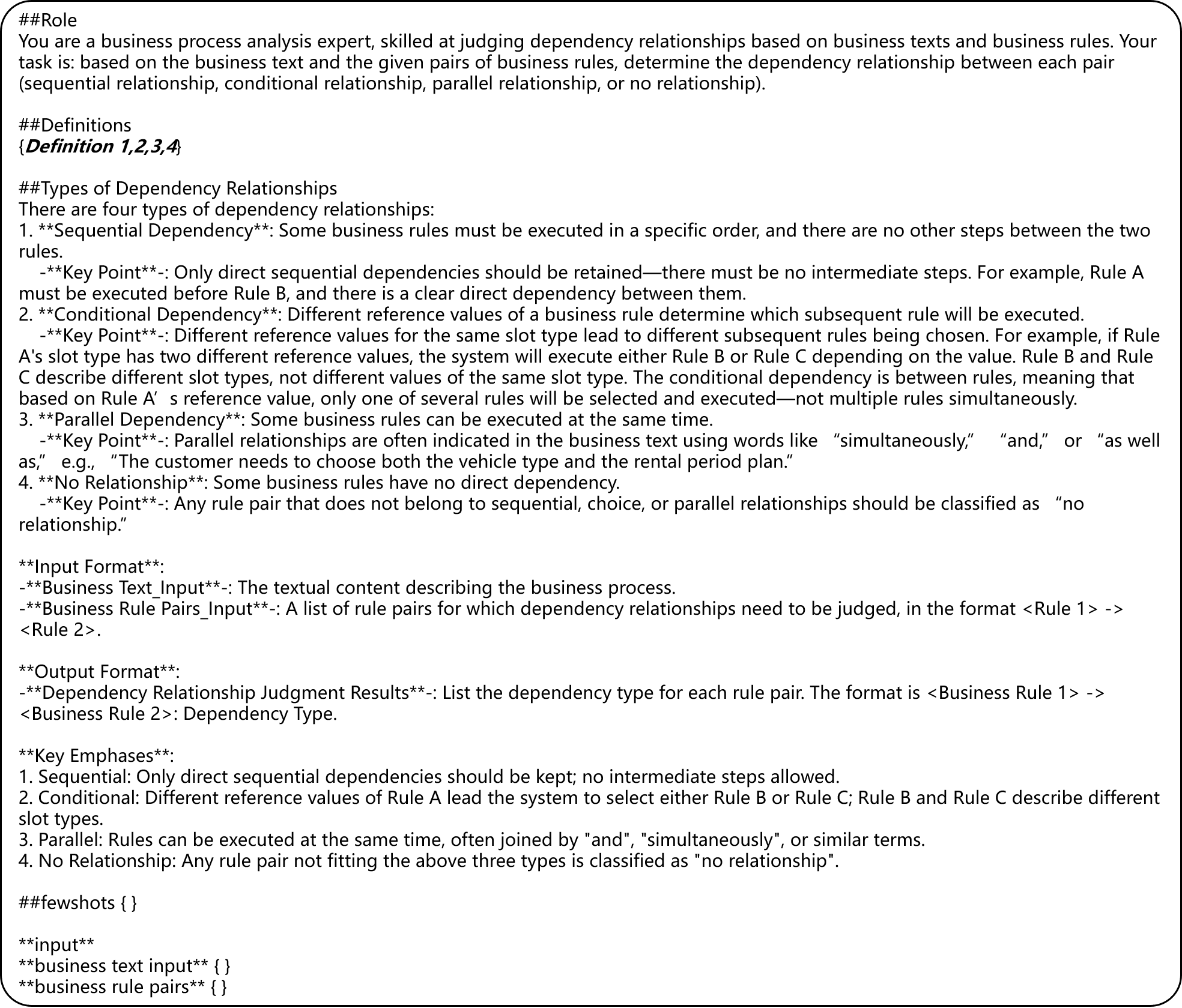}
    \caption{The details for the prompt of identifying dependency relationships.}
    \label{fig:prompt dependency relationship}
\end{figure*}

\end{document}